\documentclass{article}

\usepackage{arxiv}

\usepackage[utf8]{inputenc} 
\usepackage[T1]{fontenc}    
\usepackage{hyperref}       
\usepackage{url}            
\usepackage{booktabs}       
\usepackage{amsfonts}       
\usepackage{nicefrac}       
\usepackage{microtype}      
\usepackage{lipsum}
\usepackage{graphicx}
\graphicspath{ {./images/} }
\usepackage{authblk}
\usepackage{amsmath}
\usepackage{amssymb}
\usepackage{bm}
\usepackage{lmodern}
\usepackage{xcolor}
\newcommand{\revone}[1]{{\color{black} #1}}
\newcommand{\revtwo}[1]{{\color{black} #1}}
\newcommand{\revthree}[1]{{\color{black} #1}}
\newcommand{\editor}[1]{{\color{black} #1}}
\DeclareMathOperator{\Tr}{Tr}
\title{Exploring the Energy Landscape of RBMs: Reciprocal Space Insights into Bosons, Hierarchical Learning and Symmetry Breaking}
\author[1,2,\thanks{Corresponding author: jtoledo@triumf.ca}]{J. Quetzalc\'oatl Toledo-Marin}
\author[2]{ Anindita Maiti}
\author[3]{ Geoffrey C. Fox}
\author[4,2]{ Roger G. Melko}

\affil[1]{\footnotesize TRIUMF, Vancouver, BC V6T 2A3, Canada}
\affil[2]{\footnotesize Perimeter Institute for Theoretical Physics,
Waterloo, Ontario, N2L 2Y5, Canada}
\affil[3]{%
 University of Virginia,
 Computer Science and Biocomplexity Institute,
 994 Research Park Blvd, Charlottesville,
 Virginia, 22911, USA
}
\affil[4]{\footnotesize Department of Physics and Astronomy, University of Waterloo, Ontario, N2L 3G1, Canada}

\begin{document}
\maketitle
\begin{abstract}
Deep generative models have become ubiquitous due to their ability to learn and sample from complex distributions. Despite the proliferation of various frameworks, the relationships among these models remain largely unexplored, a gap that hinders the development of a unified theory of AI learning. In this work, we address two central challenges: clarifying the connections between different deep generative models and deepening our understanding of their learning mechanisms. We focus on Restricted Boltzmann Machines (RBMs), a class of generative models known for their universal approximation capabilities for discrete distributions. By introducing a reciprocal space formulation for RBMs, we reveal a connection between these models, diffusion processes, and systems of coupled bosons. Our analysis shows that at initialization, the RBM operates at a saddle point, where the local curvature is determined by the singular values of the weight matrix, whose distribution follows the Mar\u cenko-Pastur law and exhibits rotational symmetry. During training, this rotational symmetry is broken due to hierarchical learning, where different degrees of freedom progressively capture features at multiple levels of abstraction. This leads to a symmetry breaking in the energy landscape, reminiscent of Landau's theory. This symmetry breaking in the energy landscape is characterized by the singular values and the weight matrix eigenvector matrix. We derive the corresponding free energy in a mean-field approximation. We show that in the limit of infinite size RBM, the reciprocal variables are Gaussian distributed. Our findings indicate that in this regime, there will be some modes for which the diffusion process will not converge to the Boltzmann distribution. To illustrate our results, we trained replicas of RBMs with different hidden layer sizes using the MNIST dataset. Our findings not only bridge the gap between disparate generative frameworks but also shed light on the fundamental processes underpinning learning in deep generative models.
\end{abstract}


\section{Introduction}
Generative models are ubiquitous as they have emerged as powerful tools across multiple domains. The ongoing sprint to better models has led to a plethora of frameworks.  In data-driven contexts, these models are designed to learn and replicate the underlying probability distributions of complex datasets. \editor{ They have found significant applications in condensed matter \cite{wang2016discovering}, particularly, in quantum many-body \cite{ carleo2017solving, carrasquilla2019reconstructing,sprague2024variational}, in thermodynamics \cite{ torlai2016learning}} and nuclear physics \cite{ wang2025generalized}, where neural network-based ans\"atze are employed to model phase transitions and approximate ground-state wavefunctions. Similarly, generative models are employed in high-energy physics to address multiple problems, \textit{e.g.}, calorimeter shower generation, jet generation \cite{mikuni2022score, amram2023denoising, toledo2024conditioned} and unfolding \cite{favaro2025unfold, huetsch2025landscape}. Although generative AI techniques have achieved notable success in natural language processing and computer vision, our theoretical understanding of their learning dynamics and performance remains incomplete. Critical questions, such as the precise mechanisms through which these models capture data distributions, the robustness and fidelity of their approximations, and the criteria that determine the suitability of specific frameworks for particular datasets, remain largely unresolved. Addressing these issues is not merely an academic exercise; rather, it is essential for developing reliable and predictive generative models for complex physical systems.

Different architectures trained with different training techniques on different datasets share similar behaviors, ultimately suggesting a type of universality in generative models that goes beyond the details of each framework. In Ref. \cite{ziyin2025parametersymmetrybreakingrestoration} the authors discuss universality in deep learning in three different directions, \textit{i.e.}, the hierarchy of learning dynamics, of model complexity and of neural representation and relate these to parameter symmetry breaking. On the latter direction, it has been shown that representations of learned models are found to be universally aligned to different models trained on similar datasets \cite{kornblith2019similarity}. It then begs the question \textit{are different generative AI frameworks equivalent?} To be more precise, given the fact that different generative frameworks, \textit{e.g.}, diffusion models, restricted Boltzmann machines (RBM), generative adversarial networks (GAN) and variational auto-encoders (VAE) among others, are capable of learning the underlying distribution of the same dataset, in spite of having different architectures, training schemes and frameworks, is there an equivalence between generative AI frameworks? It is certainly the case that generative models, in general, are trained via optimizing the log-likelihood. For instance, the decoder in a VAE and the generator in a GAN both serve the role of mapping a latent representation to the dataset space. Furthermore, VAEs \cite{kingma2013auto} are trained via the evidence lower bound which is composed by a reconstruction term and the Kullback-Liebler (KL) divergence, whereas the initial GAN framework \cite{goodfellow2014generative} used the Jensen divergence for training. However, GANs are trained in an adversarial way, such that the generator learns the "likelihood" implicitly via the discriminator's feedback, while the VAE learns the likelihood explicitly.

Diffusion models were first introduced as a type of hierarchical VAE, trained by optimizing a variational bound, similar to a VAE regularizer \cite{sohl2015deep}. However, diffusion models outperform most frameworks in most tasks. Once the diffusion model is trained, the sampling process consists of first generating a Gaussian distributed random vector and then denoising it via the reverse diffusion process \cite{yu2025nonequilbrium}. This process is similar in spirit to the sampling process in an RBM\revthree{, however, diffusion models in general violate detailed balance.}

RBMs are universal approximators of discrete distributions \cite{le2008representational} and rely on computationally intensive Monte Carlo Markov Chain (MCMC) methods for training and sampling. \editor{In recent times, RBMs have gained attention within different fields in physics, in particular, from the statistical physics community \cite{decelle2017spectral, theriault2025modelling}, but also from quantum field theory \cite{aarts2024scalar}. Special interest is given to understanding the phase diagram of RBM learning \cite{hou2020statistical, bachtis2024cascade}}.
By introducing a reciprocal space formulation for RBMs, we reveal a connection between RBMs, diffusion processes, and systems of coupled Bosons. Our analysis shows that at initialization, the RBM operates at a saddle point, where the local curvature is determined by the singular values of the weight matrix, whose distribution follows the Mar\u cenko-Pastur law and exhibits rotational symmetry. During training, this rotational symmetry is 
broken due to hierarchical learning, where different degrees of freedom progressively capture features at multiple levels of abstraction. This leads to a symmetry breaking in the energy landscape, reminiscent of Landau's theory. This symmetry breaking in the energy landscape is characterized by the singular values and the weight matrix eigenvector matrix. We derive the corresponding free energy in a mean-field approximation. We show that in the limit of infinite size RBM, the reciprocal variables are Gaussian distributed. Our findings indicate that in this regime, there will be some modes for which the diffusion process will not converge to the Boltzmann distribution. We demonstrate these phenomena with an RBM and MNIST and further discuss how various initialization strategies can influence training dynamics. Our findings not only bridge the gap between disparate generative frameworks but also shed light on the fundamental processes underpinning learning in deep generative models. \revone{Additionally, recent work has demonstrated spontaneous breaking of reverse and permutation symmetries in the RBM weight‐matrix distribution at distinct critical ratios of dataset size to visible layer size \cite{hou2020statistical, theriault2025modelling}.}

The paper is organized as follows: The next section presents a brief introduction to RBMs and how they are typically trained; section \ref{sec:RBMinRS} introduces the reciprocal space, the energy landscape and the singular value distribution; section \ref{sec:PartFunc} shows the connection between RBMs with diffusion processes and systems of coupled Bosons; section \ref{sec:Sym} presents the symmetry breaking in parameter space and its connection with symmetry breaking in the energy landscape; the last section is devoted to conclusions and outlook.

\section{A brief introduction to Restricted Boltzmann Machines}
Consider a dataset $\lbrace \bm{v}^{(\alpha)} \rbrace_{\alpha=1}^{|\mathcal{D}|}$, where each data point $\lbrace 0, 1 \rbrace^{N}$ is an $N$-dimensional binary vector. Our goal is to approximate the empirical data distribution with a Boltzmann distribution, $p(\bm{v})$.
This is achieved by maximizing the log-likelihood (LL) of the model, $p(\bm{v})$, over the data set. Let us denote $P_{\mathcal{D}} = \left( \prod_{\bm{v} \in \mathcal{D}} p(\bm{v}) \right)^{1/\mathcal{D}}$. Maximizing the LL corresponds to:
\begin{equation}
    \text{argmax}_{\Theta} \ln P_{\mathcal{D}}
\end{equation}
By design, $p(\bm{v})$ is a Boltzmann distribution \textit{viz}.
\begin{equation}
    p(\bm{v}) = \frac{\sum_{\lbrace\bm{h} \rbrace} e^{-E(\bm{v},\bm{h}; \bm{a},\bm{b},\bm{W})}}{Z} \; ,
\end{equation}
where $Z = Z(\bm{a},\bm{b},\bm{W})$ is the partition function, and $E(\bm{v},\bm{h};\bm{a},\bm{b},\bm{W})$ is the energy function defined as 
\begin{equation}
    E(\bm{v},\bm{h};\bm{a},\bm{b},\bm{W}) = - \sum_i^N a_i v_i - \sum_j^M b_j h_j - \sum_{i,j} v_i W_{ij} h_j \; . \label{eq:energy}
\end{equation}
The parameters $\bm{a}$, $\bm{b}$ and $\bm{W}$ are trainable fitting parameters and $\bm{h}$ is the hidden layer, with $\bm{h} \in \lbrace 0, 1 \rbrace^{M}$. Note that the matrix $\bm{W}$ couples nodes in the visible layer, $\bm{v}$, with the nodes in the hidden layer, $\bm{h}$, and there are no explicit couplings among nodes in the visible or hidden layers, which is the same to say that the RBM is a bipartite graph (\textit{i.e.}, restricted).

To maximize the LL, we can use stochastic gradient descent. We therefore compute the gradient of the LL with respect to some generic parameter $\Theta$:
\begin{eqnarray}
    \frac{\partial \ln P_{\mathcal{D}}}{\partial \Theta} &=& \frac{1}{|D|} \sum_{\alpha=1}^{|D|} \Bigl \langle - \frac{\partial E}{\partial \Theta} \Bigr \rangle_{p(\bm{h}|\bm{v^{(\alpha)})}}
    - \Bigl \langle - \frac{\partial E}{\partial \Theta} \Bigr \rangle_{p(\bm{v},\bm{h})}
\end{eqnarray}
with
\begin{equation}
    \frac{\partial E}{\partial \Theta} = 
    \begin{cases}
        -v_k, \qquad \Theta = a_k, \\
        - h_k, \qquad \Theta = b_k, \\
        -v_k h_l, \qquad \Theta = W_{kl} \; .
    \end{cases}
\end{equation}

The LL gradient simplifies to
\begin{subequations}
\begin{align}
    \frac{\partial \ln P_{\mathcal{D}} }{\partial a_k} = \frac{1}{|\mathcal{D}|} \sum_{\alpha=1}^{|\mathcal{D}|} \langle v^{(\alpha)}_k \rangle_{p(\bm{h}|\bm{v}^{(\alpha)})} - \langle v_k \rangle_{p(\bm{v},\bm{h})} \label{eq:grad1} \\
    \frac{\partial \ln P_{\mathcal{D}}}{\partial b_k} = \frac{1}{|\mathcal{D}|} \sum_{\alpha=1}^{|\mathcal{D}|} \langle h_k \rangle_{p(\bm{h}|\bm{v}^{(\alpha)})} - \langle h_k \rangle_{p(\bm{v},\bm{h})} \label{eq:grad2} \\
    \frac{\partial \ln P_{\mathcal{D}}}{\partial W_{kl}} = \frac{1}{|\mathcal{D}|} \sum_{\alpha=1}^{|\mathcal{D}|}   \langle v^{(\alpha)}_k h_l \rangle_{p(\bm{h}|\bm{v}^{(\alpha)})} - \langle v_k h_l \rangle_{p(\bm{v},\bm{h})} \label{eq:grad3}
\end{align} \label{eq:grads}
\end{subequations}
where index $k$ and $\alpha$, respectively, refer to the RBM node and dataset point. We further have introduced the notation:
\begin{equation}
    \langle \bullet \rangle_{p(\bm{h} | \bm{v}^{(\alpha)})} = \frac{ \sum_{\lbrace\bm{h} \rbrace} \bullet \; e^{- E(\bm{v}^{(\alpha)},\bm{h})} }{ \sum_{\lbrace\bm{h} \rbrace} e^{- E(\bm{v}^{(\alpha)},\bm{h})}} \label{eq:p(h|v)}
\end{equation}
and 
\begin{equation}
    \langle \bullet \rangle_{p(\bm{v},\bm{h})} = \frac{ \sum_{\lbrace \bm{v},\bm{h} \rbrace} \bullet \; e^{-E(\bm{v},\bm{h})} }{ \sum_{\lbrace \bm{v},\bm{h} \rbrace} e^{-E(\bm{v},\bm{h})}} \; . \label{eq:p(v,h)}
\end{equation}

The first terms in Eqs. \eqref{eq:grads} can be further simplified to:
\begin{subequations}
    \begin{align}
        \frac{1}{|\mathcal{D}|} \sum_{\alpha=1}^{|\mathcal{D}|} \langle v^{(\alpha)}_k \rangle_{p(\bm{h}|\bm{v}^{(\alpha)})} &= \frac{1}{|\mathcal{D}|} \sum_{\alpha=1}^{|\mathcal{D}|} v_k^{(\alpha)} \; , \\
        \frac{1}{|\mathcal{D}|} \sum_{\alpha=1}^{|\mathcal{D}|} \langle h_k \rangle_{p(\bm{h}|\bm{v}^{(\alpha)})} &= \frac{1}{|\mathcal{D}|} \sum_{\alpha=1}^{|\mathcal{D}|} \sigma(C_k(\bm{v}^{(\alpha)})) \; , \\
        \frac{1}{|\mathcal{D}|} \sum_{\alpha=1}^{|\mathcal{D}|}   \langle v^{(\alpha)}_k h_l \rangle_{p(\bm{h}|\bm{v}^{(\alpha)})}  &= \frac{1}{|\mathcal{D}|} \sum_{\alpha=1}^{|\mathcal{D}|} v_k^{(\alpha)} \sigma(C_l(\bm{v}^{(\alpha)})) \; .
    \end{align}
\end{subequations}
Since the number of states for $\bm{v}$ and $\bm{h}$ are $2^N$ and $2^M$, respectively, the number of terms in the sums in Eq. \eqref{eq:p(v,h)} is $2^{N+M}$. This exponential dependence on the dimensionality makes computing the expectation values over $p(\bm{v},\bm{h})$ intractable for large $N$ and $M$. To overcome this limitation, importance sampling is used.
Note that $q(\bm{h}|\bm{v}) = p(\bm{v},\bm{h})/p(\bm{v})$, from which it is straightforward to show that
\begin{equation}
    q(\bm{h}|\bm{v}) = \prod_{j=1}^M q(h_j|\bm{v}) \; ,
\end{equation}
where
\begin{eqnarray}
    q(h_j|\bm{v}) = \frac{e^{h_j C_j(\bm{v})}}{1+e^{C_j(\bm{v})}} \label{eq:twoLevelEq}
\end{eqnarray}
and $C_j(\bm{v}) = \sum_i v_i W_{ij} + b_j$. Thus, the probability of node $h_j=1$ is given by $q(h_j=1|\bm{v})=\sigma(C_j(\bm{v}))$, and similarly for $v_i=1$ the probability yields $p(v_i=1 |\bm{h}) = \sigma(D_i(\bm{h}))$, with $D_i(\bm{h}) = \sum_{j} W_{ij} h_j + a_i$. The expressions $\sigma(D_i(\bm{h}))$ and $\sigma(C_j(v))$ are used for importance sampling. Note that the joint probability can be expressed as $p(\bm{v},\bm{h})=p(\bm{v}|\bm{h})p(\bm{h})$ and $p(\bm{v},\bm{h})=q(\bm{h}|\bm{v})p(\bm{v})$. Thus, we can assume a prior $p(\bm{v})$ from where we generate hidden samples $\lbrace\bm{h} \rbrace$ via $q(\bm{h}|\bm{v})$, each sample in $\lbrace\bm{h} \rbrace$ is then used to generate visible samples, $\bm{v}$. Repeating this process $K$ times yields a sequence of block Gibbs sampling steps. We denote the overall sampling process as $\lbrace \bm{v},\bm{h} \rbrace \sim \prod_{i=1}^{K} p(\bm{v}^{(i)}|\bm{h}^{(i)}) q(\bm{h}^{(i)}|\bm{v}^{(i-1)}) p(v^{(0)}) $. Here the latin index $i$ should not be mistaken with the Greek index $\alpha$ used to tag the dataset points. For large $K$, the samples can be considered to come from the stationary distribution $p(\bm{v},\bm{h})$. We repeat the importance sampling $\mathcal{N}$ times and build the estimator for the expectation value in Eq. \eqref{eq:p(v,h)} by the arithmetic average over the $\mathcal{N}$ samples generated via block Gibbs sampling:
\begin{equation}
    \langle \bullet \rangle_{p(\bm{v},\bm{h})} \approx \frac{1}{\mathcal{N}} \sum_{\lbrace \bm{v},\bm{h} \rbrace \sim \text{BGS}} \bullet  \; . \label{eq:p(v,h)2}
\end{equation}

The Gibbs sampling number of steps is commonly on the orders of $10^2$, and it has been shown that as the number of updates during training increases, the number of Gibbs sampling steps must also be increased for the RBM to reach equilibrium rather than becoming stuck in an non-equilibrium state \cite{decelle2021equilibrium}.

The standard procedure to train and RBM involves partitioning the data set $\mathcal{D}$ into mini-batches $\mathcal{D}_\chi$, such that $\mathcal{D} = \cup_\chi \mathcal{D}_\chi$ and $\cap \mathcal{D}_{\chi} = \varnothing$. The RBM parameters are then updated according to:
\begin{subequations}
\begin{align}
    a_{k}^{(t)} = a_{k}^{(t-1)} + \eta \frac{\partial \ln P_{\mathcal{D}_\chi} }{\partial a_{k}} \; , \\
    b_{k}^{(t)} = b_{k}^{(t-1)} + \eta \frac{\partial \ln P_{\mathcal{D}_\chi} }{\partial b_{k}} \; , \\
    W_{kl}^{(t)} = W_{kl}^{(t-1)} + \eta \frac{\partial \ln P_{\mathcal{D}_\chi} }{\partial W_{kl}} \; , \label{eq:wandbUpdates}
\end{align}
\end{subequations}

where $\eta$ is the learning rate. Three primary training strategies are discussed in the literature, each differing mainly on how the Markov chain in Eq. \eqref{eq:p(v,h)2} is initialized. \textit{Rdm-K}: For each parameter update, the initial $\bm{h}$ is randomly sampled from a Bernoulli distribution. Here $K$ is the number of block Gibbs sampling steps. \textit{Contrastive Divergence}: In CD, a data point from the training dataset is used as the initial $\bm{v}$ vector for the block Gibbs sampling. This method improves performance in RBMs compared to Rdm-K \cite{decelle2021equilibrium}. \textit{Persistent contrastive divergence}: This method is similar to CD where the Markov chain is started using a data point in the data set for the first parameter update, while for the remaining parameter updates, the Markov chains are initialized using the last state in the previous parameter update. Although training with PCD can be challenging, when executed correctly it achieves superior results compared to CD \cite{fernandez2023disentangling}. Additional techniques, such as centering the RBM gradient and incorporating regularizers, have also been proposed to enhance performance \cite{hinton2010practical, cho2011enhanced, melchior2016center}

\begin{figure}[hbtp]
\centering
\includegraphics[width=3.4in]{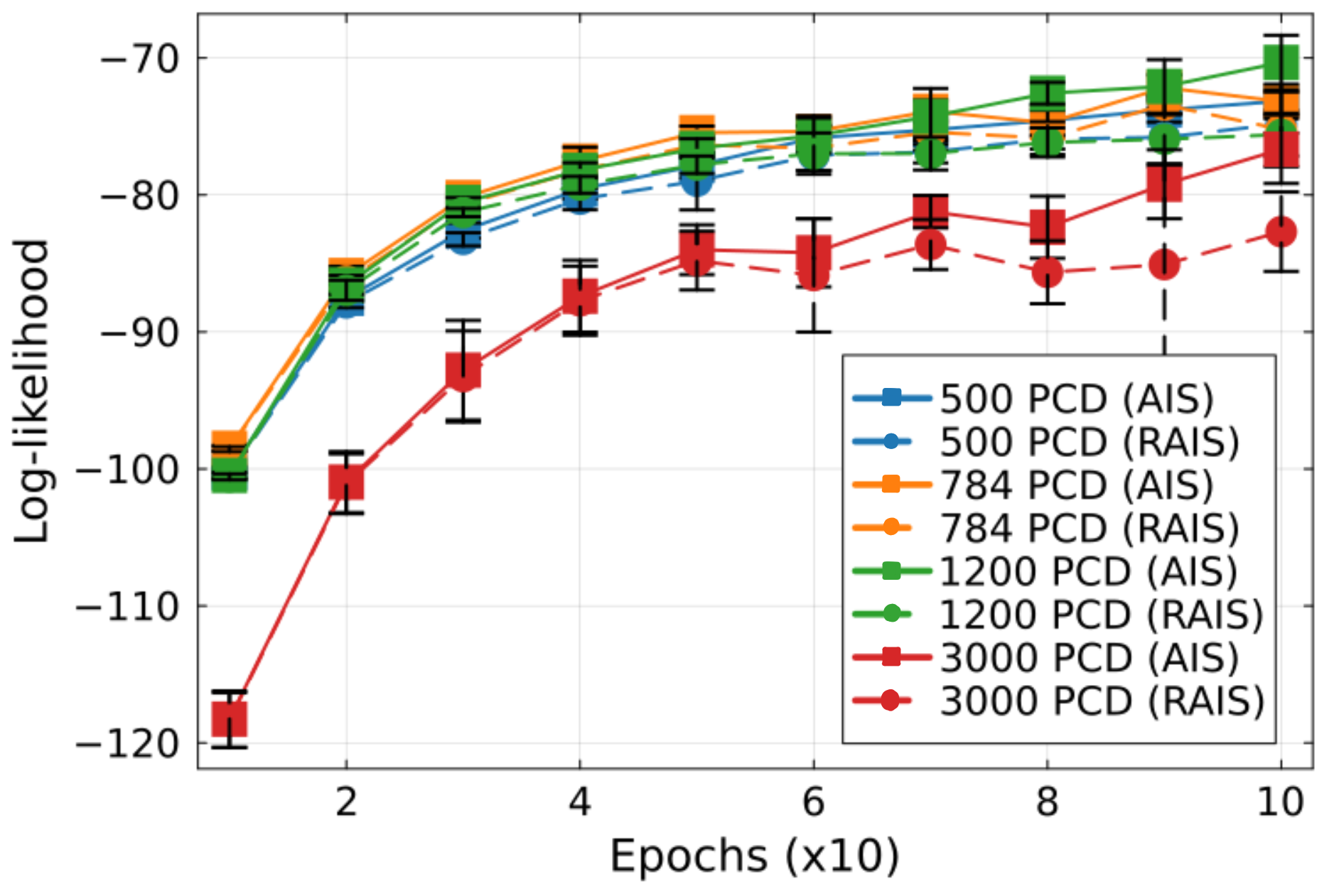}
\includegraphics[width=3.0in]{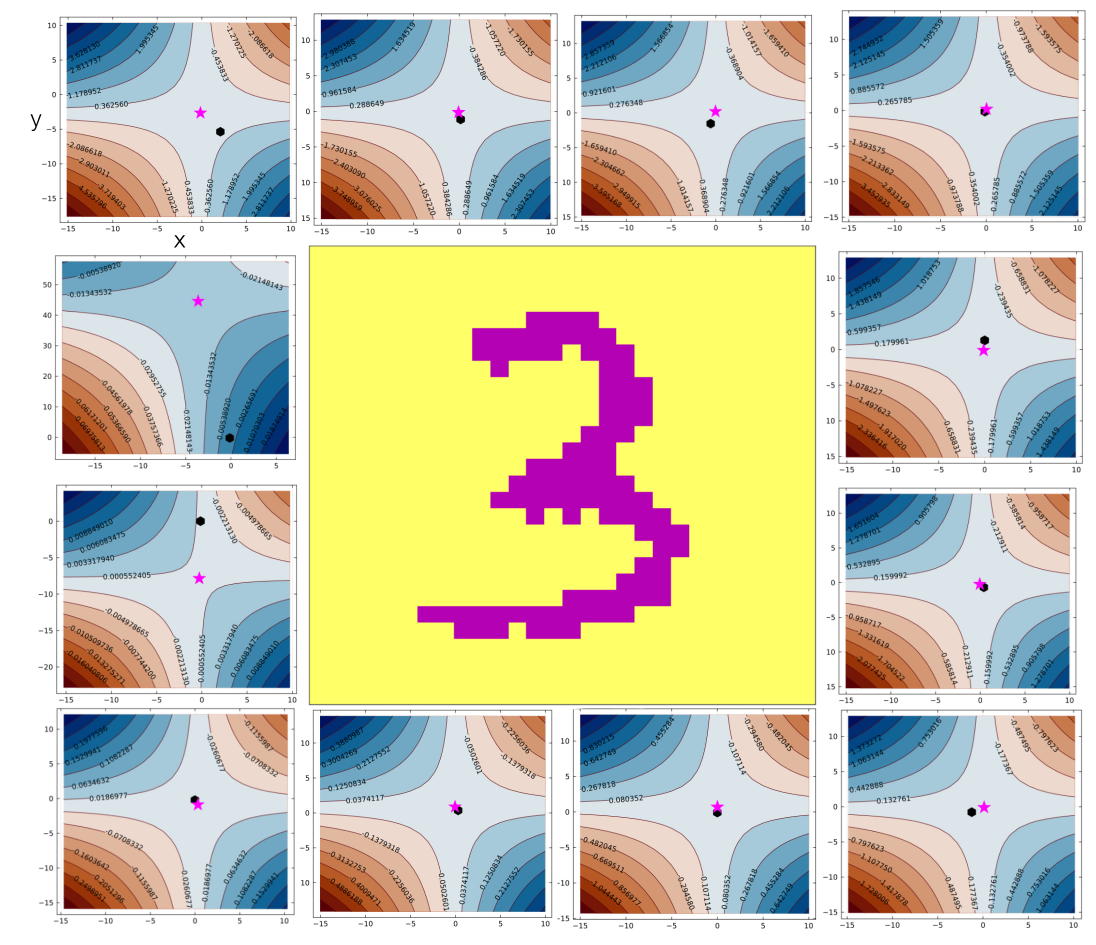}
\caption{\textbf{Left panel)} Log-likelihood \textit{vs} epochs for RBMs with hidden layer sizes $M=500,784,1200,3000$ . The partition function was estimated using annealed importance sampling (AIS) and reverse AIS \cite{salakhutdinov2008learning, burda2015accurate}. Each data point corresponds to the average over five replicas and the error bars correspond to the standard deviation. \textbf{Right panel)} Image of number $3$ generated from a trained RBM and, via the transformations in Eq. \eqref{eq:transformations}, projected onto the energy landscapes (clockwise starting at the upper left corner) $1,2,3,4,20,30,50,100,200,300,498$ and $500$. The magenta star marks the saddle point whereas the black pentagon corresponds to the image projection to reciprocal space.} \label{fig:pcdReciprocalSpaceNumberThree}
\end{figure}

All the numerical results regarding trained RBMs presented in this paper were obtained by training on \editor{MNIST dataset \cite{lecun1998mnist, JuliaPkg}} and are available in our Github repository \cite{Toledo-Marin_RBM}. The RBM training was done using the publicly available Julia code, as described in Ref. \cite{fernandez2023disentangling}. We trained five replicas for each RBM with hidden layer size $500,784,1200$ and $3000$. In Fig. \ref{fig:pcdReciprocalSpaceNumberThree} left panel we show the log-likelihood \textit{vs} epochs to verify the maximization of the log-likelihood. We estimated the partition function using annealed importance sampling and reverse annealed importance sampling, which correspond to an upper and lower bound of the partition function \cite{salakhutdinov2008learning, burda2015accurate}. 

\section{Restricted Boltzmann machine in reciprocal space} \label{sec:RBMinRS}

Updating the weight matrix and the self-fields via Eqs. \eqref{eq:wandbUpdates} ultimately reduces the energy associated to the dataset points via the energy function (see Eq. \eqref{eq:energy}). The energy landscape described by the energy function has at most as many saddle points as there are units in the smaller partition. At initialization, the dataset points cluster around these saddle points; however, as training progresses, they deviate from the saddle points but never drift away, due to implicit constraint imposed by the binary nature of the variables. To show this, we project the RBM onto reciprocal space \revone{by projecting the state vectors via the eigenvectors of weight matrix} using singular value decomposition (SVD). \revone{Projecting the system onto this reciprocal space, we obtain a new set of variables with single pairwise interaction each. These new variables capture collective patterns weighed by the eigenvectors}. By studying the eigenvector matrices and singular values of the weight matrix, we learn about the weight matrix properties and provide a physical interpretation of the singular values in the context of RBMs. \revthree{We highlight that in Refs. \cite{decelle2017spectral, decelle2018thermodynamics} a similar approach has been proposed as means to study the learning dynamics in RBMs. In particular, the authors study the learning dynamics in terms of the singular values of the weight matrix.}

\revtwo{Upon initialization}, the coupling matrix $\bm{W}$ is a rectangular, non-symmetric random matrix with Gaussian entries and standard deviation $\sigma$. We perform SVD on $\bm{W}$, such that $\bm{W} = \bm{U} \bm{\Sigma} \bm{V}^t = \sum_{\alpha} \lambda_{\alpha} | v_{\alpha} \rangle \langle h_\alpha |$; where $\bm{U}$ is a matrix composed of the eigenvectors of $\bm{W} \cdot \bm{W}^t$, \textit{i.e.}, $\bm{U} = \sum_s | v_s \rangle \langle c_s |$, where  $ \lbrace | c_s \rangle \rbrace$ denotes the canonical basis; $\Sigma$ is a rectangular matrix of size $N \times M$ containing the singular values $\lbrace \lambda_i \rbrace_{i=1}^{min(N,M)}$ on the diagonal; and $\bm{V}^t$ is a matrix composed of the eigenvectors of $\bm{W}^t \cdot \bm{W}$, \textit{i.e.}, $\bm{V}^t = \sum_s | c_s \rangle \langle h_s |$. Note that due to the orthogonality of $\bm{U}$ and $\bm{V}$, we have $\bm{U}\cdot \bm{U}^t = \bm{V} \cdot \bm{V}^t = \bm{I}$. Moreover, the eigenvector matrices $\bm{U}$ and $\bm{V}$ are Haar distributed, \textit{i.e.}, they are uniformly distributed over the group $O(N)$ and each eigenvector is uniformly distributed on the unit hypersphere \cite{potters2020first}. The training process breaks this symmetry, as we will show later.

Using the matrices $\bm{U}$ and $\bm{V}^t$, we can express the energy function, Eq. \eqref{eq:energy}, as:
\begin{eqnarray}
    E(\bm{v},\bm{h}) \rightarrow E(\bm{x},\bm{y}) =   - \langle x | a_0 \rangle - \langle b_0| y \rangle - \langle x | \bm{\Sigma} | y \rangle  \; . \label{eq:energy_xy}
\end{eqnarray}
Expressing the energy in terms of the singular value decomposition of the coupling matrix recasts the model in terms of quasi-discrete variables with single pairwise interactions. We say quasi-discrete variable because each $x_i$ ($y_j$) can take on values formed by summing subsets of elements in column $i$-th ($j$-th) of matrix $U$ ($V$). The single pairwise interactions allows us to decompose the energy as a sum of individual pairwise energy terms:
\begin{equation}
    E(\bm{x},\bm{y}) = \sum_i E_i(x_i,y_i) \; , \label{eq:energyInxy}
\end{equation}
such that
\begin{equation}
    E_i(x_i,y_i) = \begin{cases}
        - a_{0i} x_i - b_{0i} y_i - \lambda_{i} x_i y_i, \quad 1 \leq i \leq min(N,M) \\
        - a_{0i} x_i, \quad min(N,M) < i \leq max(N,M) \\
    \end{cases}
\end{equation}
where
\begin{subequations}
\begin{align}
    \begin{cases}
        \bm{a_0} = \bm{U}^t \bm{a}  \\ 
        \bm{b_0} = \bm{V}^t \bm{b} \\ 
        \bm{x} = \bm{U}^t \bm{v}  \\
        \bm{y} = \bm{V}^t \bm{h} 
    \end{cases} \label{eq:transformations}
\end{align}
\end{subequations}
 \revtwo{ It is instructive to consider a dynamical system described by the potential $E_i(x_i,y_i)$, such that the state of the system is given by $x_i$ and $y_i$ and a velocity field $(\dot{x}_{i}, \dot{y}_{i}) = ( \partial E_i(x_i,y_i)/\partial x_i, \partial E_i(x_i,y_i)/\partial y_i)$ \cite{percival1982introduction}. The dynamics of the system are described by the set of equations:
\begin{equation}
    \begin{pmatrix}
        \dot{x}_{i} \\
        \dot{y}_{i}
    \end{pmatrix}
    =-
    \begin{pmatrix}
        a_{0i} \\
        b_{0i}
    \end{pmatrix}
    -
    \begin{pmatrix}
        0 & \lambda \\
        \lambda & 0 
    \end{pmatrix}
    \begin{pmatrix}
        x_{i} \\
        y_{i}
    \end{pmatrix}
\end{equation}

Setting the velocity field to zero yields the fixed point of the system.} From the Hessian matrix we find that the eigenvalues correspond to $\pm \lambda_i$, indicating that the fixed point is a saddle point. Each of the $E_{i(\leq min(N,M))}$ corresponds to a hyperbolic paraboloid with saddle point at $(x_{0i},y_{0i})=(-b_{0i}/\lambda_i, -a_{0i}/\lambda_i)$ where the energy is equal to $b_{0i} a_{0i}/\lambda_i$. In other words, for each energy landscape $E_{i(\leq min(N,M))}$ the phase curves are such that there is always a direction where the energy decreases and tends to $-\infty$. \revtwo{Another way to see this is by noticing that the energy landscape $E_i(x_i,y_i)$ is the equation of a conic section, in particular, of a hyperbola.  }

In Fig. \ref{fig:pcdReciprocalSpaceNumberThree} we show the projection of an image of number $3$ generated from a trained RBM using the transformations in Eq. \eqref{eq:transformations}. The energy landscapes are ordered clockwise starting from the upper left corner for indices $i=1,2,3,4,20,30,50,100,200,300,498$ and $500$. The magenta star marks the saddle point, whereas the black pentagon corresponds to the image projection to reciprocal space.

\begin{figure}[hbtp]
\centering
\includegraphics[width=3.0in]{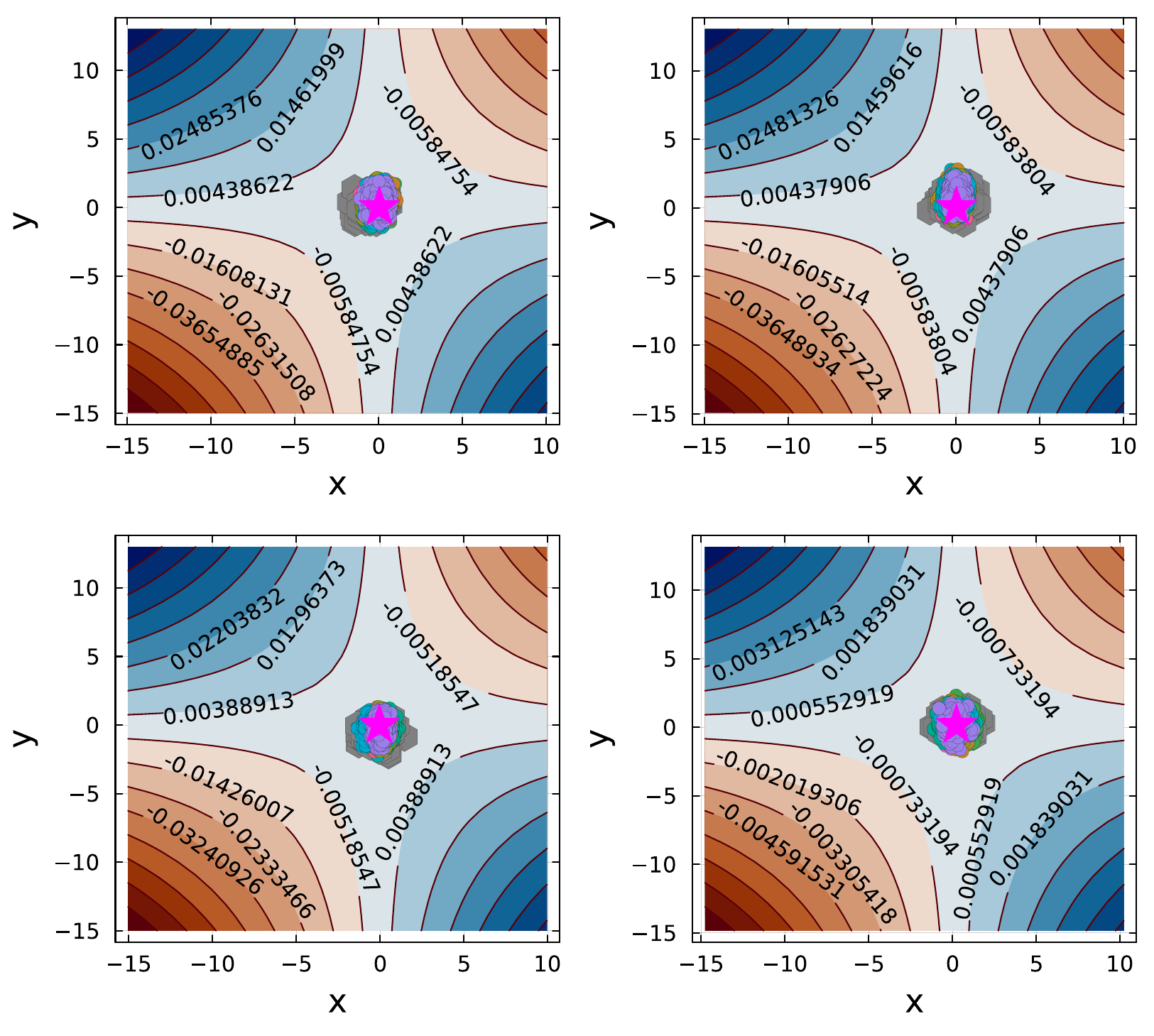}
\includegraphics[width=3.0in]{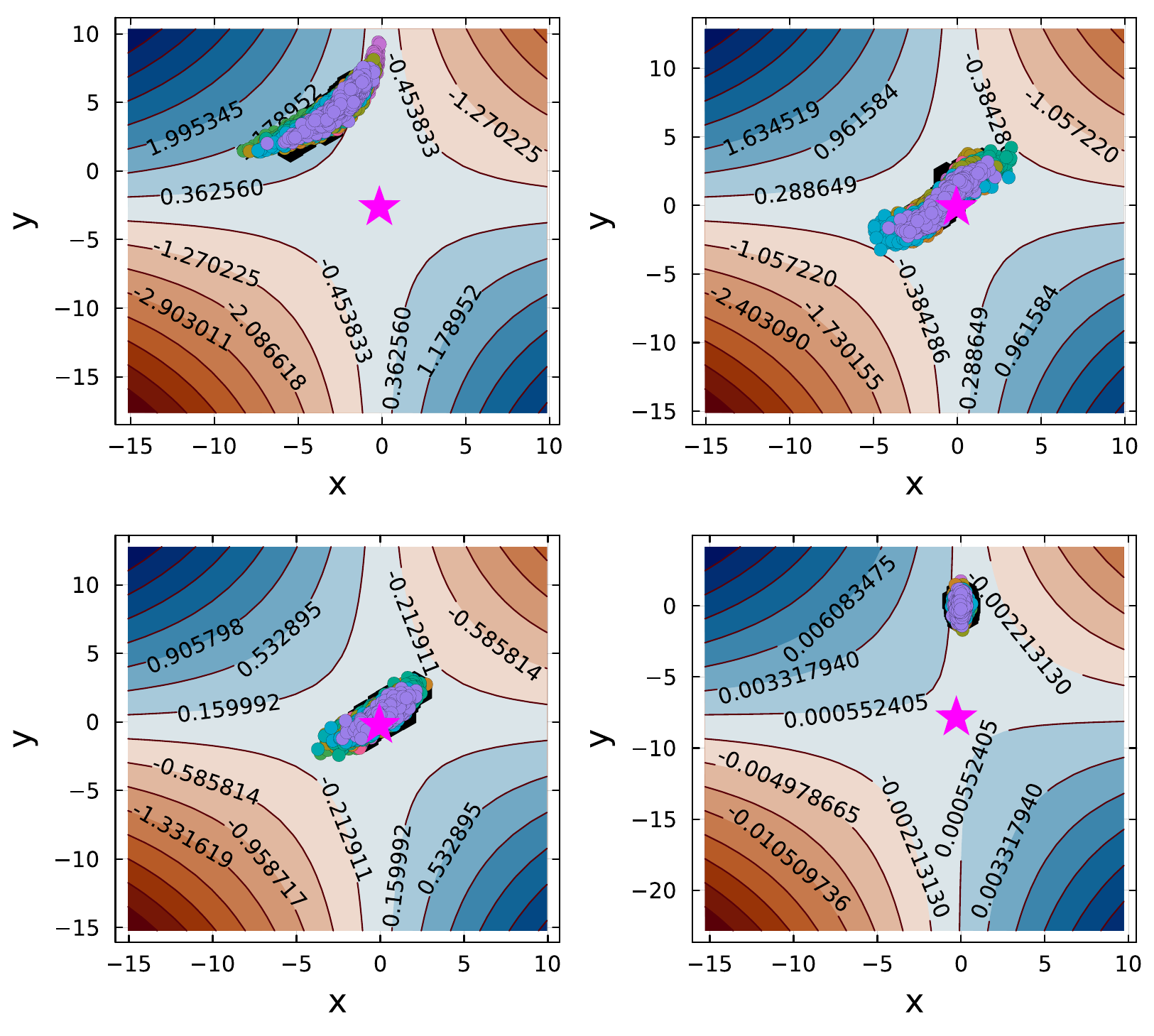}
\caption{ Each subpanel shows the $xy$-plane contour energy $E_i$ for the singular value index 1,2,3 and 498. The colored data points correspond to MNIST test data projected onto the reciprocal space. The gray hexagons correspond to Gibbs sampled data. The magenta star marks the saddle point. \textbf{Left panel)} Randomly initialized RBM. \textbf{Right panel)} Trained RBM.} \label{fig:pcdReciprocalSpace}
\end{figure}

\revtwo{We define a set of variables} $\lbrace u_i \rbrace_{i=1}^{\text{min(N,M)}},\lbrace w_i\rbrace_{i=1}^{\text{min(N,M)}}$ which relate to $x_i,y_i$ for $1 \leq i \leq min(N,M)$ via a translation to the saddle points and a rotation, namely,
\begin{equation} \label{eq:uiwi}
    \begin{pmatrix}
        x_i \\
        y_i 
    \end{pmatrix} =
    \begin{pmatrix}
        x_{0i} \\
        y_{0i} 
    \end{pmatrix} +
    \begin{pmatrix}
        \cos \theta & -\sin \theta \\
        \sin \theta & \cos \theta 
    \end{pmatrix}
    \begin{pmatrix}
        u_{i} \\
        w_{i} 
    \end{pmatrix} 
\end{equation}
Setting $\theta=\pi/4$ aligns the principal axis of the hyperbolic paraboloid with the $uw$-plane axis, leading to the energy function having the following form:
\begin{eqnarray}
    E_i(u_i,w_i) = \frac{a_{0i}b_{0i}}{\lambda_i} - \frac{\lambda_i}{2}(u_i^2 - w_i^2) \; .
\end{eqnarray}
We can now give a physical interpretation of the weight matrix singular values. The singular values can be interpreted as the vibrational modes, since the curvature of each mode per energy landscape is $\pm \lambda$. In addition to modulating the stiffness of the mode, the singular values modulate the saddle point energy and its location. In Fig. \ref{fig:pcdReciprocalSpace} we show four energy landscapes in the $xy$-plane corresponding to the singular value indices $1,2,3$ and $498$ in the case of a trained RBM, and we also project the test set onto these landscapes. We do this for an untrained and a trained RBM. In the former, the dataset points cluster on the saddle point, whereas for the trained RBM this is not always the case although the points do not drift away from the saddle point.

\begin{figure}
    \centering
    \includegraphics[width=6.4in]{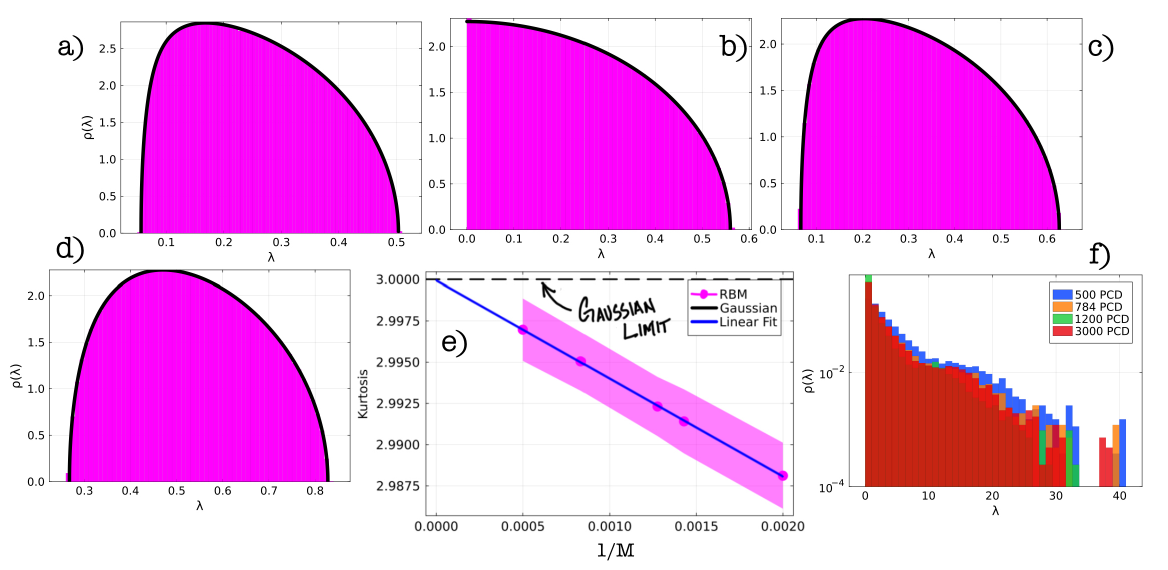}
    \caption{Singular values probability density function for a random RBM with 784 visible nodes and with \textbf{a)} 500, \textbf{b)} 784, \textbf{c)} 1200 and \textbf{d)} 3000 hidden nodes. \textbf{e)} Kurtosis of the reciprocal variable $y$ for hidden layer sizes $M=500,700,784,1200,2000$ for randomly initialized RBMs, with $784$ visible nodes. Each data point was generated from averaging over the kurtosis of the hidden nodes with non-zero singular values. The kurtosis for each node was computed from projecting $7\cdot 10^6$ binary vector samples to reciprocal space. The ribbon corresponds to the standard deviation over the hidden layer. The dashed line marks the target for Gaussianity. The blue curve is a linear fit extrapolated to hidden layer size infinity where the reciprocal variable becomes Gaussian. The linear fit residue is on the order of $\mathcal{O}(10^{-4})$ and has been included. \textbf{f)} Distribution of singular values for trained RBM with different hidden layer sizes and different training methods.}
    \label{fig:rhoLambda}
\end{figure}

In practice, several empirical guidelines are used for initializing RBMs, many of which are supported by experimental evidence. For example, selecting the number of nodes in the hidden layer remains largely heuristic. A common recommendation is to limit the hidden layer to no more nodes than the visible layer, as exceeding this threshold may slow down training due to the increase in the number of operations per matrix multiplication. An additional reason to not have the hidden layer be much greater the the visible nodes is related to the singular values distribution. As we have shown previously, the singular values influence the stability of the energy landscape. \revthree{As noted previously in Refs. \cite{decelle2017spectral, decelle2018thermodynamics}}, the singular value spectrum is characterized by the Mar\u cenko–Pastur law \cite{potters2020first}:
\begin{equation}
    \varrho(\lambda) = \frac{\sqrt{(\lambda_+^2 - \lambda^2)(\lambda^2 - \lambda_-^2)}}{\pi q \lambda} \qquad \lambda_-<\lambda<\lambda_+ \; , \label{eq:MP}
\end{equation}
with $\lambda_{\pm} = \sqrt{\text{max}(N,M) \sigma^2}\pm\sqrt{q}$ and $q=\text{min}(N,M)\sigma^2$. In Fig. \ref{fig:rhoLambda} we plot Eq. \eqref{eq:MP} for various RBM sizes. One can demonstrate that the maximum of $\varrho(\lambda)$ occurs at $\lambda_{max}=\sqrt{\lambda_+ \lambda_-}$, implying $\lambda_{max} = \sigma\sqrt{|N-M|}$. This distribution increases with the difference between visible and hidden nodes as well as with increasing standard deviation $\sigma$. Consequently, a significant imbalance between the numbers of visible and hidden nodes and/or a broad range of coupling values accentuates the singular value distribution, thereby impacting the stability of the energy landscapes at initialization. \revtwo{Note that if the weight matrix is initialized with $\sigma^2 \sim 1/\text{max}(N,M)$, the extensive scaling with the dimensions drops out and only a dependence on $\text{max}(N,M)/\text{max}(N,M)\leq1$ remains.}

It is interesting to note that while the energy is minimized as $(x_i,y_i)\rightarrow\pm( \infty, \infty)$, the dataset clusters around the saddle point, suggesting an implicit constraint during training. This property ultimately comes from the constraints imposed by the binary space. In the next section we obtain these constraints as a potential in reciprocal space.

\section{Partition function in reciprocal space} \label{sec:PartFunc}
In the previous section we presented the RBM in reciprocal space, we gave a physical interpretation of the singular values and we described the general structure of the energy landscape of the RBM. We also mentioned that the implicit constraints imposed by the binary nature of the variables confine the dataset points to regions near the saddle points. In this section we treat the reciprocal variables as continuous while enforcing the constraints imposed by the binary nature of the variables as a potential acting on these degrees of freedom. This approach allows us to map an RBM to a diffusion process and, subsequently, to a Schr\"odinger equation describing the interaction of Bosons. 

We first construct the partition function of the reciprocal space in terms of the variables $\lbrace u_i \rbrace_{i=1}^{\text{min(N,M)}},\lbrace w_i\rbrace_{i=1}^{\text{min(N,M)}}$ and $\lbrace x_j \rbrace_{j=\text{min}(N,M)+1}^{\text{max}(N,M)}$. We follow a similar path to the replica method in the case of a single replica. Applying the replica method to RBMs has been done previously in \revthree{Refs.} \cite{decelle2018thermodynamics, decelle2021restricted}.

To simplify the notation let us assume that $M<N$.
Let us consider $u_{1:M}$, $w_{1:M}$ and $x_{M+1:N}$ continuous variables subject to the constraints given by Eq. \eqref{eq:transformations} \textit{viz}.,

\begin{eqnarray}
    Z = 2^{N+M} \int du_{1:M} dw_{1:M} dx_{M+1:N} e^{-\beta E(\bm{u},\bm{w}, \bm{x})} \rho(\bm{u},\bm{w},\bm{x}) \label{eq:partition_uwx}
\end{eqnarray}
with
\begin{eqnarray}
    \rho(\bm{u},\bm{w},\bm{x}) &\equiv& \langle \prod_{i=1}^M \prod_{j=M+1}^N \delta(u_i - \frac{1}{\sqrt{2}}(\sum_k U_{ik}^t v_k - \sum_k V_{ik}^t h_k - x_{0i} + y_{0i})) \nonumber \\
    && \delta(w_i - \frac{1}{\sqrt{2}}(\sum_k U_{ik}^t v_k + \sum_k V_{ik}^t h_k - x_{0i} - y_{0i})) \delta(\sum_{k} U_{jk}^t v_k - x_j)  \rangle_{(\bm{v},\bm{h})} \nonumber \\
    &=& \frac{1}{2^{N+M}} \sum_{\lbrace \bm{v},\bm{h} \rbrace} \prod_{i=1}^M \prod_{j=M+1}^N \delta(u_i - \frac{1}{\sqrt{2}}(\sum_k U_{ik}^t v_k - \sum_k V_{ik}^t h_k - x_{0i} + y_{0i})) \nonumber \\
    && \delta(w_i - \frac{1}{\sqrt{2}}(\sum_k U_{ik}^t v_k + \sum_k V_{ik}^t h_k - x_{0i} - y_{0i})) \delta(\sum_{k} U_{jk}^t v_k - x_j)
\end{eqnarray}
By expressing the partition function as in \eqref{eq:partition_uwx}, the intractable term is isolated within the density function $\rho(\bm{u},\bm{w}, \bm{x})$. This density function encodes the constraints imposed on the continuous variables $\bm{u}$, $\bm{w}$ and $\bm{x}$ by the discreteness of the binary variables $\bm{v}$ and $\bm{h}$. These constraints appear as an effective potential in the continuous variables phase space. In a similar manner to the replica method, we can express the density function $\rho(\bm{u},\bm{w}, \bm{x})$ in Fourier space. After some straightforward algebra, we find that:
\begin{eqnarray}
    Z = \int du_{1:M} d\hat{u}_{1:M} dw_{1:M} d\hat{w}_{1:M} dx_{M+1:N} d\hat{x}_{M+1:N} e^{\mathcal{S}} \label{eq:partition_uwx_FT}
\end{eqnarray}
where the variables $\hat{u}_{1:M}$, $\hat{w}_{1:M}$ and $\hat{x}_{M+1:N}$ are the conjugate variables and $\mathcal{S}$ is the effective action:
\begin{eqnarray}
    \mathcal{S} = -\beta E(\bm{u},\bm{w},\bm{x}) + i \left( \sum_i ( \hat{u}_i u_i + \hat{w}_i w_i ) + \sum_j \hat{x}_j x_j \right) - (N+M)\ln 2\pi \\
    + \ln \Tr_{\lbrace \bm{v},\bm{h} \rbrace } e^{-i \sum_i (\hat{u}_i f_i(\bm{v},\bm{h}) + \hat{w}_i g_i(\bm{v},\bm{h}) ) - i \sum_j \hat{x}_j \sum_k U_{jk}^t v_k} 
\end{eqnarray}
The saddle point approximation leads to the following equalities:
\begin{subequations}
\begin{align}
    \hat{u_l}^* &= i \beta \lambda_l u_l \\
    \hat{w_l}^* &= - i \beta \lambda_l w_l \\
    \hat{x_l}^* &= i \beta a_{0l} \\
    u_l^* &= \langle \langle f_i(\bm{v},\bm{h}) \rangle \rangle \equiv \Bigl \langle \Bigl \langle \frac{1}{\sqrt{2}}(\sum_k U_{lk}^t v_k + \sum_k V_{lk}^t h_k - x_{0l} - y_{0l}) \Bigr \rangle \Bigr \rangle \\
    w_l^* &= \langle \langle g_i(\bm{v},\bm{h}) \rangle \rangle \equiv \Bigl \langle \Bigl \langle \frac{1}{\sqrt{2}}(- \sum_k U_{lk}^t v_k + \sum_k V_{lk}^t h_k + x_{0l} - y_{0l}) \Bigr \rangle \Bigr \rangle \\
    x_l^* &= \Bigl \langle \Bigl \langle \sum_k U_{lk}^t v_k \Bigr \rangle \Bigr \rangle
\end{align}
\end{subequations}
Here we have introduced the auxiliary function $f_i(\bm{v},\bm{h})$ and $ g_i(\bm{v},\bm{h})$ and we have denoted the expectation value
\begin{equation}
    \langle \langle \bullet \rangle \rangle = \frac{\Tr_{\lbrace \bm{v},\bm{h} \rbrace } \bullet e^{-i \sum_i (\hat{u}_i f_i(\bm{v},\bm{h}) + \hat{w}_i g_i(\bm{v},\bm{h}) ) - i \sum_j \hat{x}_j \sum_k U_{jk}^t v_k}}{\Tr_{\lbrace \bm{v},\bm{h} \rbrace } e^{-i \sum_i (\hat{u}_i f_i(\bm{v},\bm{h}) + \hat{w}_i g_i(\bm{v},\bm{h}) ) - i \sum_j \hat{x}_j \sum_k U_{jk}^t v_k}}
\end{equation}
The solution $u^*_{1:M}$, $w^*_{1:M}$, $x^*_{M+1:N}$, $\hat{u}^*_{1:M}$, $\hat{w}^*_{1:M}$ and $\hat{x}^*_{M+1:N}$ corresponds to the saddle point. In principle, one can solve these equations self-consistently similar to the replica method \cite{decelle2021restricted}. Note that in the high temperature limit the conjugate variables tend to zero in the saddle point approximation while the variables $u^*_{1:M}$, $w^*_{1:M}$ and $x^*_{M+1:N}$ tend to the arithmetic average over the binary states $\bm{v}$ and $\bm{h}$:
\begin{subequations}
    \begin{align}
        u^{*}_{l \; (T \gg \lambda_l)} &= \frac{1}{\sqrt{2}} \left( \sum_k U_{lk}^t \langle v_k \rangle + \sum_k V_{lk}^t \langle h_k \rangle - x_{0l} - y_{0l} \right) \\
        w^{*}_{l \; (T \gg \lambda_l)} &= \frac{1}{\sqrt{2}} \left( - \sum_k U_{lk}^t \langle v_k \rangle + \sum_k V_{lk}^t \langle h_k \rangle + x_{0l} - y_{0l} \right) \\
        x^*_{l \; (T \gg \lambda_l)} &= \sum_k U_{lk}^t \langle v_k \rangle
    \end{align} \label{eq:saddlePoint}
\end{subequations}
In this context, any binary state has an equal probability $2^{-N-M}$, which implies that $\langle v_k \rangle = \langle h_k \rangle = 1/2$ for all $k=1,...,N+M$.

We mentioned in the previous section that $x_i$ ($y_j$) can take on values formed by summing subsets of elements in column $i$-th ($j$-th) of matrix $\bm{U}$ ($\bm{V}$). Additionally, in a randomly initialized RBMs the eigenvector matrices $\bm{U}$ and $\bm{V}$ are Haar distributed. Hence the clustering of data around the saddle point can be understood as a consequence of the Haar measure and the central limit theorem. The distribution of $u_{1:M}$, $w_{1:M}$ and $x_{M+1:N}$ is close to Gaussian. In Fig. \ref{fig:rhoLambda} e) we show the kurtosis of the reciprocal variable $y$ for hidden layer sizes $500,700,784,1200,2000$ for randomly initialized RBMs, with $784$ visible nodes. Each data point was generated from averaging over the kurtosis of the hidden nodes with non-zero singular values. The kurtosis for each node was computed from projecting $7\cdot 10^6$ random binary vector samples to reciprocal space. The ribbon corresponds to the standard deviation over the hidden layer. The dashed line marks the target for Gaussianity. The blue curve is a linear fit extrapolated to hidden layer size infinity where the reciprocal variables become Gaussian. In the case of a trained RBM, the variables $u_{1:M}$, $w_{1:M}$ and $x_{M+1:N}$ are not close to Gaussian, in general.
However, for both trained and untrained RBMs, the mean of $u_l$ is equal to $u^*_{l \; (T \gg \lambda_l)}$ and the same applies to $w_l$ and $x_l$ as we will show later. The results in Eqs. \eqref{eq:saddlePoint} will be useful for when we expand the constraint potential when solving the Schr\"odinger equation as well as for when we introduce the mean field to show the symmetry breaking in reciprocal space.

In the next section we present the connection with the Fokker-Planck Equation and a set of coupled Bosons. We anticipate further connections between RBMs with diffusion-like models and coupled Bosons.

\subsection{Fokker-Planck Equation and coupled Bosons} \label{sec:FP}
In the previous section we expressed the partition function in terms of the continuous variables $u_{1:M}, w_{1:M}$ and $x_{M+1:N}$, under the assumption that $M<N$. To make notation homogeneous, we introduce the variable $\bm{z} = \left[u_{1:M}, w_{1:M}, x_{M+1:N} \right]^T$.
The partition function becomes:
\begin{equation}
    Z = \int d\bm{z} e^{-\beta U_{eff}(\bm{z})}
\end{equation}
where 
\begin{equation}
    U_{eff}(\bm{z}) = E(\bm{z}) - \frac{S_c}{\beta}+\frac{V_{const}(\bm{z})}{\beta} \label{eq:effectivepotential}
\end{equation}
with $S_c$ denoting the configurational entropy $(N+M)\ln 2$, and 
\begin{align}
V_{const}(\bm{z})=-\ln \rho(\bm{z}) 
\end{align} 
is the constraint potential that arises due to the constraints imposed via the original discrete variables. We assume the potential $U_{eff}(\bm{z})$ is a confining potential, such that the probability density function
\begin{equation}
    P(\bm{z}) = \frac{e^{-\beta U_{eff}(\bm{z})}}{Z} \label{eq:PDF_stat}
\end{equation}
\revtwo{In this sense, we now have mapped the RBM distribution to a Boltzmann distribution of a set of continuous variables described by the effective potential $U_{eff}(\bm{z})$ immersed in a heat bath with inverse temperature $\beta$. Therefore, Eq \eqref{eq:PDF_stat} is the stationary solution of the Fokker-Planck equation \cite{risken1996fokker}:}
\begin{equation}
    \frac{\partial P(\bm{z},t)}{\partial t} = D \sum_{i=1}^{N+M}\left[ \frac{\partial^2 P(\bm{z},t)}{\partial z_i^2} + \frac{1}{D} \frac{\partial  }{\partial z_i} \left( \frac{\partial U_{eff}(\bm{z})}{\partial z_i} P(\bm{z},t)\right)\right] \; . \label{eq:FPeq}
\end{equation}
We assume the over-damped regime with the friction coefficient set to $\gamma=1$ such that \revtwo{ the diffusion coefficient, defined via the fluctuation-dissipation theorem, becomes $D=1/\beta$. We also highlight that it may be possible to recast the training dynamics under a framework of Dyson Brownian dynamics as done in Ref. \cite{aarts2024dyson}}. It is straightforward to show that Eq. \eqref{eq:PDF_stat} is the stationary solution to Eq. \eqref{eq:FPeq}. By the variable separation method we propose the ansatz $P(\bm{z},t)=\phi(\bm{z})f(t)$, which substituting in Eq. \eqref{eq:FPeq} and dividing by the ansatz leads to:
\begin{equation}
    \frac{\dot{f}(t)}{f(t)} = \frac{D}{\phi} \sum_{i=1}^{N+M}\left[ \frac{\partial^2 \phi(\bm{z})}{\partial z_i^2} + \frac{1}{D} \frac{\partial  }{\partial z_i} \left( \frac{\partial U_{eff}(\bm{z})}{\partial z_i} \phi(\bm{z})\right)\right] = - \Gamma
\end{equation}
Solving for $f(t)$ yields $f(t) = A e^{-\Gamma t}$, where $A$ is a constant and $\Gamma >0$ in order for the solution to not diverge. The parameter $\Gamma$ is the inverse of the characteristic relaxation time.

There are many ways to solve the remaining of the Fokker-Planck Equation. Here we reduce the equation to a time-independent Schr\"odinger-like equation by introducing the substitution:
\begin{equation} \label{eq:phifokkerplanck}
    \phi(\bm{z}) = e^{-U_{eff}(\bm{z})/2D} \psi(\bm{z}) \; .
\end{equation}
This leads to an eigenvalue problem $H\psi(\bm{z}) = E \psi(\bm{z})$, where $H=\sum_i^{N+M}\frac{p_i^2}{2} + V_Q^{(i)}(\bm{z})$ ($m=\hbar=1$). Here $p_i$ is the momentum operator and $V_Q$ is the potential defined as:
\begin{equation}
    V_Q^{(i)}(\bm{z}) = \frac{1}{8D^2}\left(\frac{\partial U_{eff}(\bm{z})}{\partial z_i}\right)^2 - \frac{1}{4D} \frac{\partial^2 U_{eff}(\bm{z})}{\partial z_i^2}
\end{equation}

We can formally express the general solution as:
\begin{equation}
    P(\bm{z},t)= \sum_{n} c_{\Gamma_n}e^{-\frac{1}{2D}U_{eff}(\bm{z})} \psi_{\Gamma_n}(\bm{z}) e^{-\Gamma_n t}
\end{equation}
with $\Gamma_n = E_n/2D$. The ground state corresponds to $\Gamma_0=E_0/2D=0$, such that we can rewrite the previous general solution as:
\begin{equation}
    P(\bm{z},t)= c_0 e^{-\frac{1}{2D}U_{eff}(\bm{z})} \psi_{0}(\bm{z})  + \sum_{\Gamma_n>0} c_{\Gamma_n}e^{-\frac{1}{2D}U_{eff}(\bm{z})} \psi_{\Gamma_n}(\bm{z}) e^{-\Gamma_n t} \label{eq:GeneralSolutionSchr}
\end{equation}
from which it is easy to notice that $c_0=1/Z$ and $\psi_0(\bm{z})=e^{-\frac{1}{2D}U_{eff}(\bm{z})}$.

The explicit general solution will depend on the effective potential $U_{eff}(\bm{z})$. In general, we can consider expanding the effective potential around a minimum.
Notice that the maximum log-likelihood of the point PDF $\rho(\bm{z})$ corresponds to the minimum in $V_{const}(\bm{z})$. Hence, we can approximate the constraint potential by Taylor expansion around the minimum, 
\begin{equation}
    V_{const}(\bm{z}) \approx V_{const}(\bm{\mu}_z) + \frac{1}{2}\sum_{ij} (z_i - \mu_{z_i}) k_{ij} (z_j - \mu_{z_j}) \; , 
\end{equation}
where $k_{ij}=\frac{\partial^2 V_{const}(\bm{z})}{\partial z_i \partial z_j}|_{\bm{\mu}_z}$ and $\mu_z$ is the expectation value of $z$, which correspond to the high temperature limit in the saddle point expansion (see Eqs. \eqref{eq:saddlePoint}). Before moving forward, we introduce the parameters
\begin{equation}
    \omega_i =
    \begin{cases}
        \frac{k_{ii} - \lambda \beta}{2} \quad \text{for }i=1,...,M \\
        \frac{k_{ii} + \lambda \beta}{2} \quad \text{for }i=M+1,...,2M \\
        \frac{k_{ii}}{2}  \quad \text{for }i=2M+1,...,N \\
    \end{cases}
\end{equation}
and
\begin{equation}
    z_{i0} =
    \begin{cases}
        \frac{\mu_{z_i}k_{ii}}{2\omega_i} \quad \text{for }i=1,...,2M \\
        \frac{\beta a_i}{2\omega_i} + \mu_{z_i}  \quad \text{for }i=2M+1,...,N \\
    \end{cases}
\end{equation}
in addition to making the change in variable $\zeta_i = z_i - z_{i0}$.
After some algebra, the Schr\"odinger potential becomes:
\begin{eqnarray}
    V_Q^{(i)} &=& \frac{1}{2} \omega_i^2 \zeta_i^2 + \frac{1}{2}\omega_i \zeta_i\sum_{j\neq i}\zeta_j k_{ij} + \frac{1}{2}\omega_i \zeta_i \sum_{j\neq i}(z_{j0} - \mu_{z_j}) k_{ij} \nonumber \\
    && + \frac{1}{2} \left(\frac{1}{2}\sum_{j \neq i} (\zeta_j + z_{j0} - \mu_{z_j}) k_{ij} \right)^2 - \frac{1}{2}\omega_i + \mathcal{O}(\zeta^4)
\end{eqnarray}
The previous potential correspond to that of $N+M$ coupled harmonic oscillators. We can rewrite the potential $V_Q(\bm{\zeta})$ as:
\begin{equation}
    V_Q(\bm{\zeta}) \approx \frac{1}{2} \langle \bm{\zeta} | M | \bm{\zeta} \rangle + \frac{1}{2} \langle \bm{\zeta} | O | \bm{\Delta} \rangle + \frac{1}{2} \langle \bm{\Delta} | \Theta | \bm{\Delta} \rangle - \frac{1}{2} \langle \bm{1} | \Omega | \bm{1} \rangle \; . \label{eq:Q_pot}
\end{equation}
where
\begin{equation}
    \begin{cases}
        \Delta_k = z_{k0}-\mu_{z_k} \\
        \Theta_{ij}=\frac{1}{4} \sum_{k \neq j, k \neq i} k_{ik} k_{kj} \\
        M_{ij} = \omega^2_i \delta_{ij} + w_i k_{ij} (1-\delta_{ij}) + \Theta_{ij} \\
        O_{ij} = \omega_i k_{ij} (1-\delta_{ij}) + 2 \Theta_{ij} \\
        \Omega_{ij} = \omega_i \delta_{ij}
    \end{cases}
\end{equation}
The matrix $M$ is a square symmetric matrix, which can be diagonalized.

We can obtain $k_{ij}$ by assuming that the reciprocal variables are Gaussian distributed, \textit{viz.}
\begin{equation}
    \rho(\bm{z}) = \frac{\sqrt{det(\bm{k})}}{(2\pi)^{N/2}} e^{- \frac{1}{2} \sum_{ij} (z_i - \mu_{z_i}) k_{ij} (z_j - \mu_{z_j})}
\end{equation}
which leads to
\begin{equation}
    \frac{1}{k_{ij}} = \langle (z_i - \mu_{z_i})(z_j - \mu_{z_j}) \rangle = \frac{1}{4}\delta_{ij}
\end{equation}
The previous implies that $\Theta_{ij}=O_{ij}=0$ for all $i$ and $j$. The potential in Eq. \eqref{eq:Q_pot} reduces to:
\begin{equation}
    V_{0}^{(i)} = \frac{1}{2}\omega_i^2 \zeta_i^2 - \frac{1}{2} \omega_i
\end{equation}
Notice that in this case, the problem reduces to a set of $N+M$ uncoupled harmonic oscillators with a shift in the potential energy. The eigenstates to the Schr\"odinger Equation reduce to a product of Hermite polynomials, while the eigenvalues are
\begin{equation}
    E(\bm{n}) = \sum_{i=1}^{N+M} \omega_i \left(n_i + \frac{1}{2}\right) - \frac{\omega_i}{2} = \sum_{i=1}^{N+M} \omega_i n_i \label{eq:dispersion}
\end{equation}
where $n_i=0,1,2,...$ for all $i$. The inverse characteristic relaxation time becomes $\Gamma(\bm{n})=\beta E(\bm{n})/2$. Notice that when $\bm{n}=\bm{0}$ then $E(\bm{0})=0$ corresponding to the stationary solution, as expected. For $\bm{n} \neq \bm{0}$, when $\lambda_i > k_i/\beta$ the eigenvalues become negative and the solution diverges. In practice, we do not observe this behavior since the oscillators are in general coupled. Therefore, higher-order terms in the constraint potential need to be considered. In a previous section we showed that the reciprocal variables are Gaussian distributed when either of the partition sizes tend to infinity (see Fig. \ref{fig:rhoLambda} e). In such scenario, we expect the constraint potential to have terms up to second order. Consequently, there will be some modes for which the diffusion process will not converge to the Boltzmann distribution (see Eq. \eqref{eq:GeneralSolutionSchr}).

Note that the only approximation here was the Taylor expansion of the constraint potential, which ultimately is a standard approach when solving a many-body problem. In this sense, we have shown a clear connection between RBM, diffusion processes and coupled Bosons.

\section{Symmetry breaking} \label{sec:Sym}
Deep learning process occurs hierarchically, \textit{i.e.}, different degrees of freedom capture different features of the data. These features usually have a hierarchical sorting \cite{sclocchi2025phase}. The hierarchical learning process is accompanied by a symmetry breaking and restoration in parameter space \cite{ziyin2025parametersymmetrybreakingrestoration}. In the case of RBMs, it has been observed that the eigenvector with the highest singular value aligns with the principal components of the dataset \cite{decelle2021restricted}. This process can be understood as hierarchical learning. Here we show that training the RBM breaks the rotational symmetry in parameter space. In addition, this symmetry breaking is tied to the hierarchical learning. We show this by randomly rotating the eigenvector matrix and observing the effect it has on the image class. Lastly, we also show a symmetry breaking in the Landau sense in the energy landscape during training.

\begin{figure}
    \centering
    \includegraphics[width=6.4in]{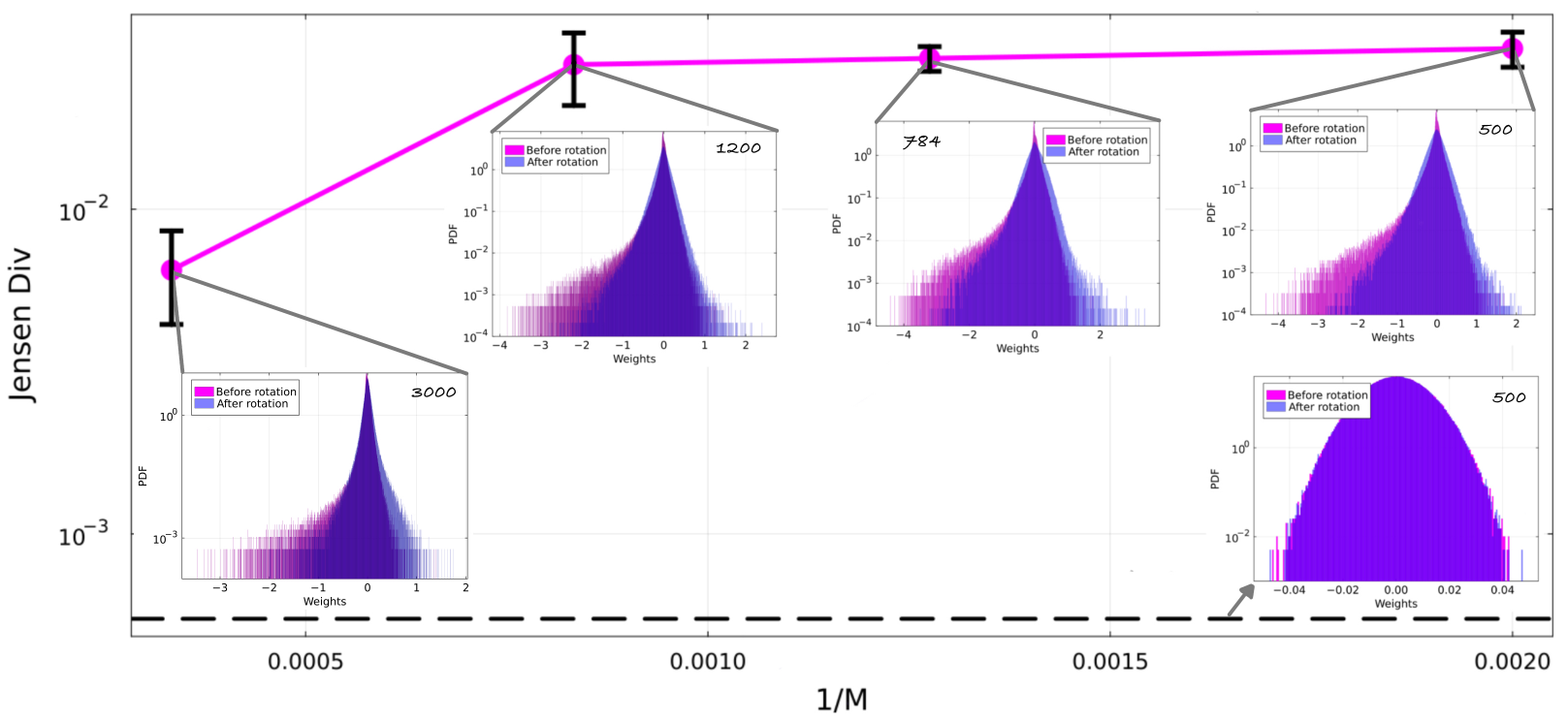}
    \caption{Jensen divergence between rotated and non-rotated weight matrix for RBMs with $M=500,784,1200,3000$. Each point corresponds to an average over five replicas. The dashed line corresponds to a non-trained RBM with $M=500$. PDF of RBM weight matrix before and after random rotations for one of the replicas is shown.}
    \label{fig:rand_rot_weights}
\end{figure}

\subsection{Rotational Symmetry}

\begin{figure}
    \centering
    \includegraphics[width=4.7in]{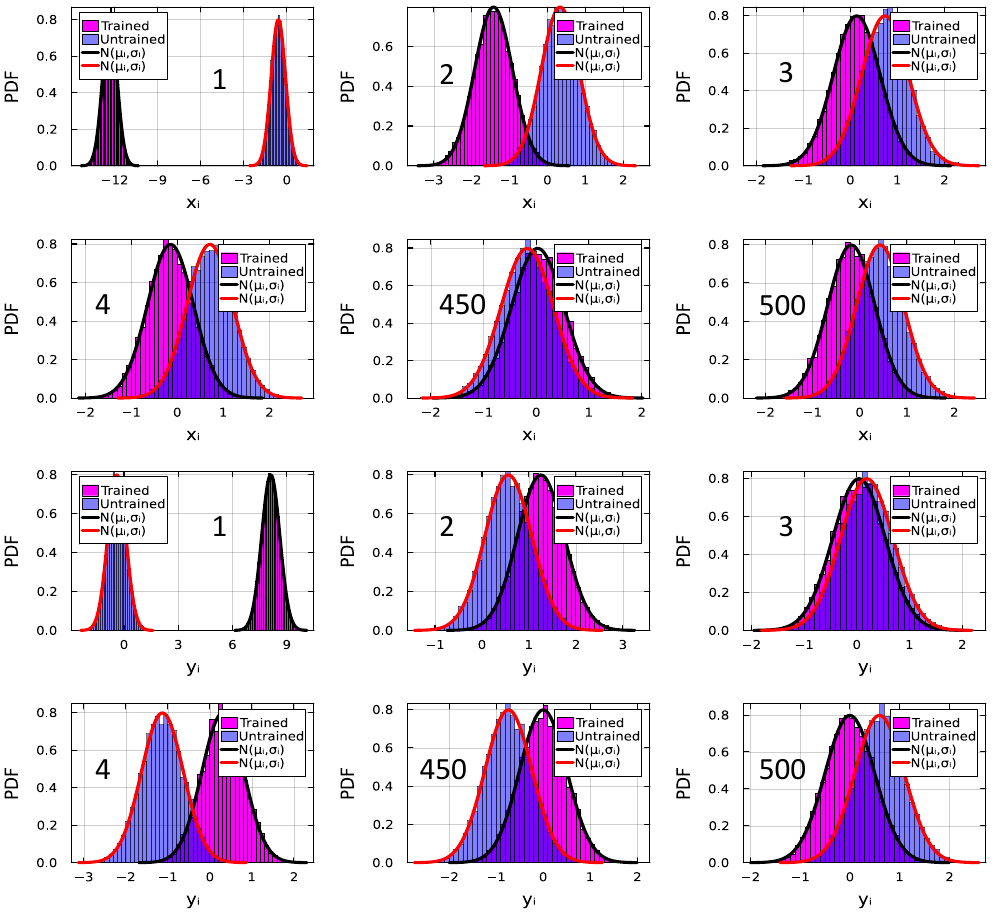}
    \caption{Probability density function of the reciprocal variables for various energy landscapes for trained and non-trained RBM. The samples where generated from a multivariate $1/2$-Bernoulli distribution projected onto reciprocal space. The continuous curves correspond to a normal distribution centered at $\mu_x^{(i)}$ and $\mu_y^{(i)}$ and standard deviation $\sigma_x^{(i)}$ and $\sigma_y^{(i)}$ given by \eqref{eq:muAndSigma}.}
    \label{fig:DistributionInRecSpace}
\end{figure}

RBMs are typically initialized by drawing each element of the weight matrix independently from a Gaussian distribution with mean $0$ and standard deviation $0.01$ \cite{hinton2012practical}. This initialization ensures that the RBM begins in a paramagnetic phase \cite{de1978stability, decelle2021restricted}. Moreover, this procedure inherently confers rotational invariance to the joint probability density of the weight elements. In fact, the joint probability density of the weight elements is given by
\begin{equation}
    p(\lbrace W_{ij} \rbrace ) = \left( \frac{1}{\sqrt{2\pi \sigma^2}} \right)^{NM} e^{-\sum_{i,j} \frac{W_{ij}^2}{2\sigma^2}} \propto \exp \left( - \Tr (\bm{W}\bm{W}^t) \right)
\end{equation}
Since $\bm{W}\bm{W}^t$ can be decomposed as $\bm{W}\bm{W}^t = \bm{U}\bm{\Sigma} \bm{\Sigma}^t \bm{U}^t$, consider a rotation matrix $\mathcal{R}$, and define the rotated orthogonal matrix $\bm{U}_R = \mathcal{R} \bm{U} \mathcal{R}^t$. Noting that 
\begin{equation}
    \Tr \left(\bm{U}_R \bm{\Sigma} \bm{\Sigma}^t \bm{U}_R^t \right)=\Tr \left(\bm{U} \bm{\Sigma} \bm{\Sigma}^t \bm{U}^t \right)=\Tr \left(\bm{\Sigma} \bm{\Sigma}^t \right) = \sum_i \lambda^2_i \; ,
\end{equation}
it follows that $\Tr (\bm{W}\bm{W}^t)$ is also rotationally invariant and a similar argument holds for $\Tr (\bm{W}^t\bm{W})$. Therefore, $p(\lbrace W_{ij} \rbrace )$ is rotationally invariant.

We probe the rotational symmetry in trained RBMs by applying random rotations on the eigenvector matrices $\bm{U}$ and $\bm{V}$, and studying the effect on the weight matrix distribution. We build a rotation matrix using the method described in Ref. \cite{aguilera2004general}. This method performs rotations in $n-$dimensions around any arbitrary $(n-2)$-dimension subspace. We then apply $N*0.1$ and $M*0.1$ consecutive random rotations to random subspaces of $(N-2)$-dimension and $(M-2)$-dimension of $\bm{U}$ and $\bm{V}$, respectively. We then compare the weight matrix before and after the random rotations in the case of an initialized RBM with 500 hidden nodes, and trained RBMs with $500, 784, 1200$ and $3000$ hidden nodes by plotting the weight matrix PDF in Fig. \ref{fig:rand_rot_weights}. We also show the Jensen divergence between the rotated and non-rotated weight matrix. Each point corresponds to 5 replicas. The black dashed line corresponds to a non-trained RBM. This shows that the training process breaks the rotational symmetry. \revone{In addition to rotational symmetry, reverse symmetry likewise breaks during training. Moreover, it is easy to note that hierarchical learning is incompatible with permute symmetry, which ultimately breaks during training, as shown in \cite{hou2020statistical, theriault2025modelling}.}

Rotational invariance of the joint distribution ensures that the eigenvector matrices $\bm{U}$ and $\bm{V}$ are Haar distributed, \textit{i.e.}, they are uniformly distributed over the group $O(N)$ and each eigenvector is uniformly distributed on the unit hypersphere \cite{potters2020first}. This property enables the use of the central limit theorem by noticing that projecting binary states to reciprocal space corresponds to summing row elements of $\bm{U}^t$ and $\bm{V}^t$ from random positions. But even for trained RBMs where the rotational symmetry is broken, the behavior of projection of binary states onto reciprocal space shows centrality. To illustrate this, we randomly sampled $10^4$ states and projected each into the reciprocal space for both, a randomly initialized RBM and a trained RBM. In Fig. \ref{fig:DistributionInRecSpace} we show the resulting probability density functions for various energy landscape indices. The continuous curves correspond to a normal distribution centered at $\mu_x^{(i)}$ and $\mu_y^{(i)}$ and standard deviation $\sigma_x^{(i)}$ and $\sigma_y^{(i)}$, defined as
\begin{subequations}
    \begin{align}
        \mu_x^{(i)} = \frac{1}{2}\sum_{j}U_{ij}^t \; , \\
        \mu_y^{(i)} = \frac{1}{2}\sum_{j}V_{ij}^t\; , \\
        \sigma_x^{(i)} = \sigma_y^{(i)} = \frac{1}{2} \; ,
    \end{align} \label{eq:muAndSigma}
\end{subequations}
which is straightforward to obtain from computing the first and second moment of the reciprocal variables. Notice that the first moment in the previous Equation is equivalent to the saddle point expansion at high temperature obtained in Eqs. \eqref{eq:saddlePoint}. Moreover, these expectation values were used in section \ref{sec:FP} for the Taylor expansion. In the following, we show that the rotational symmetry breaking in parameter space is tied to hierarchical learning in the RBM.

\begin{figure}
    \centering
    \includegraphics[width=2.2in]{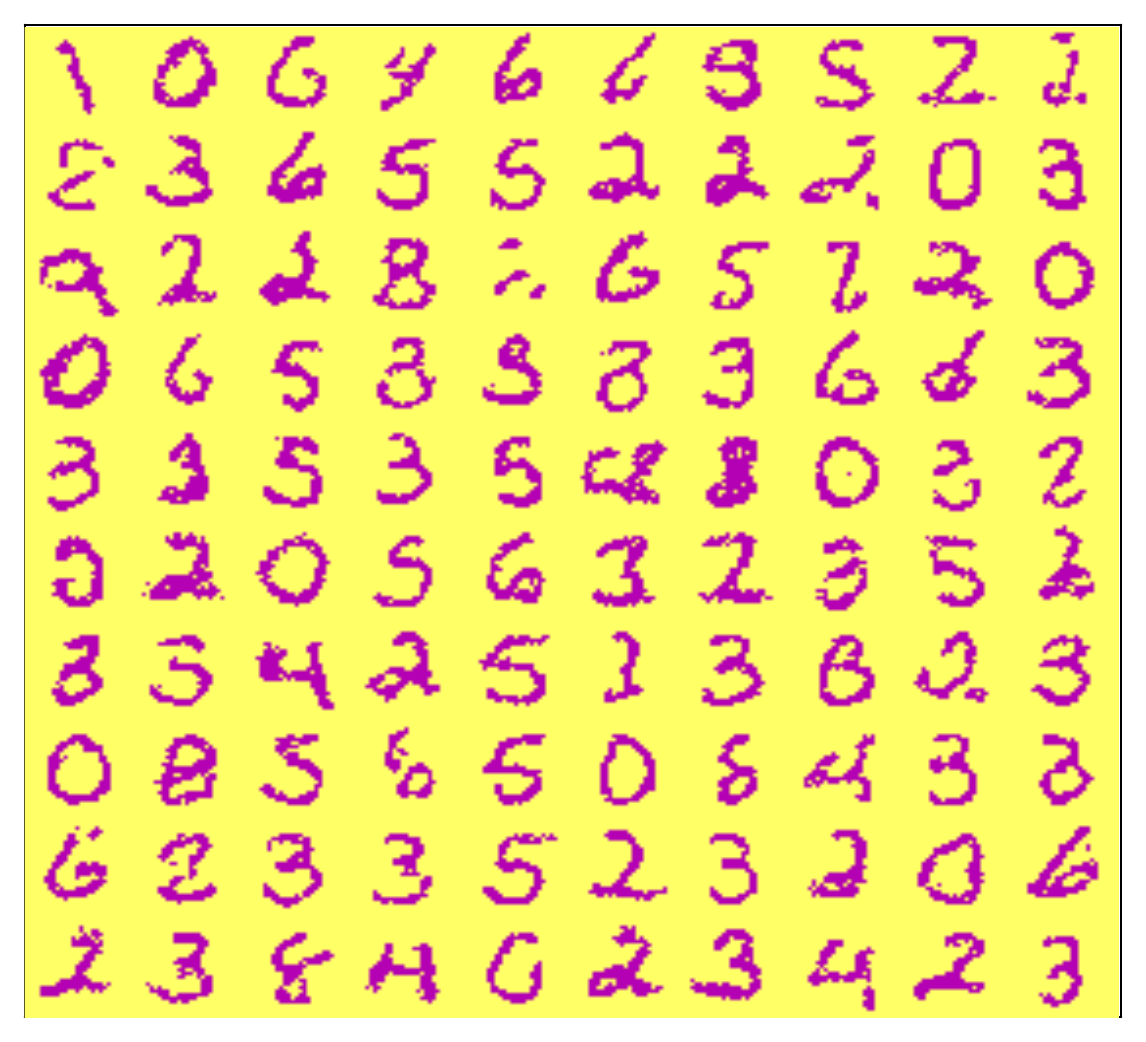}
    \includegraphics[width=2.2in]{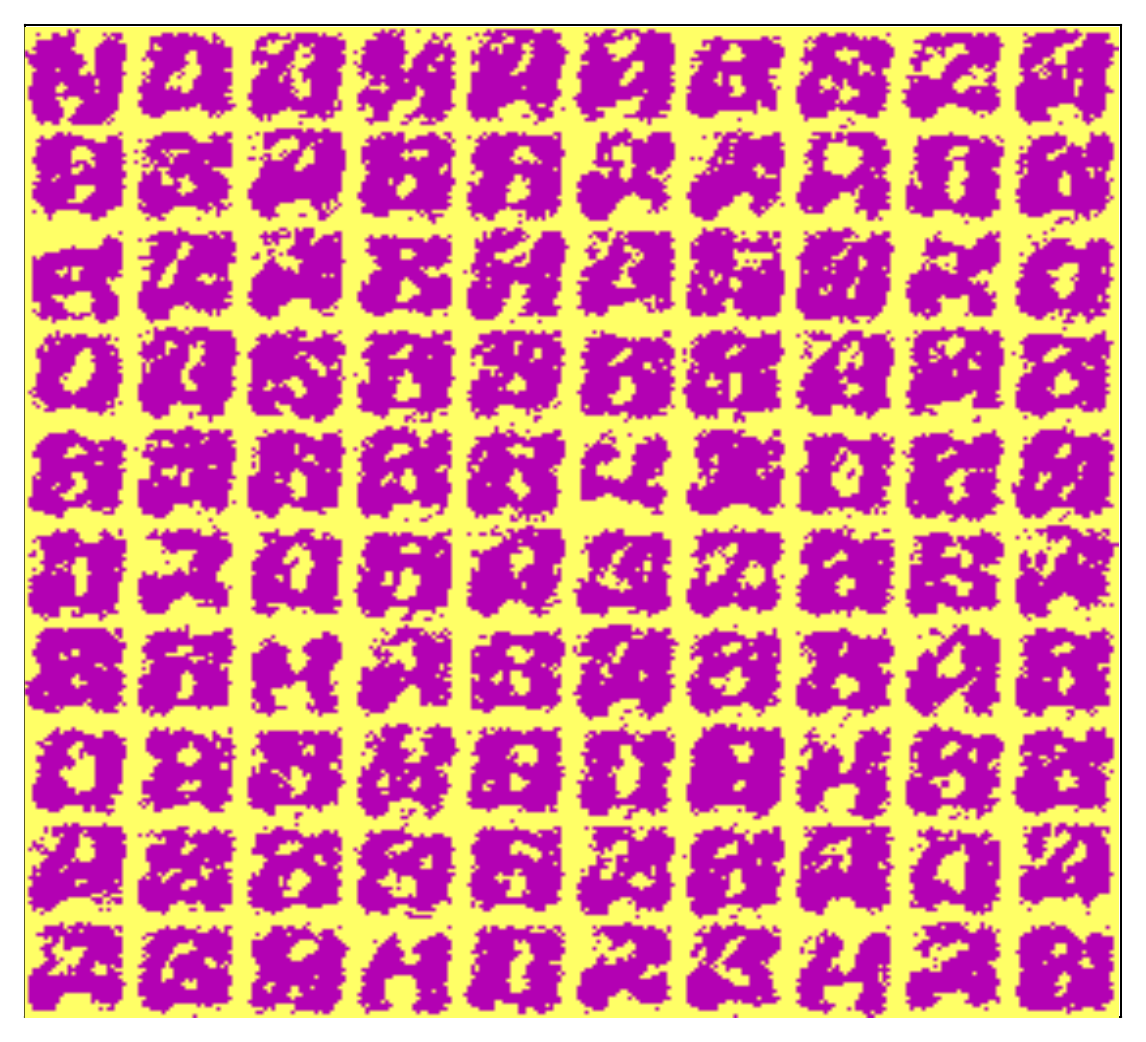}
    \includegraphics[width=2.2in]{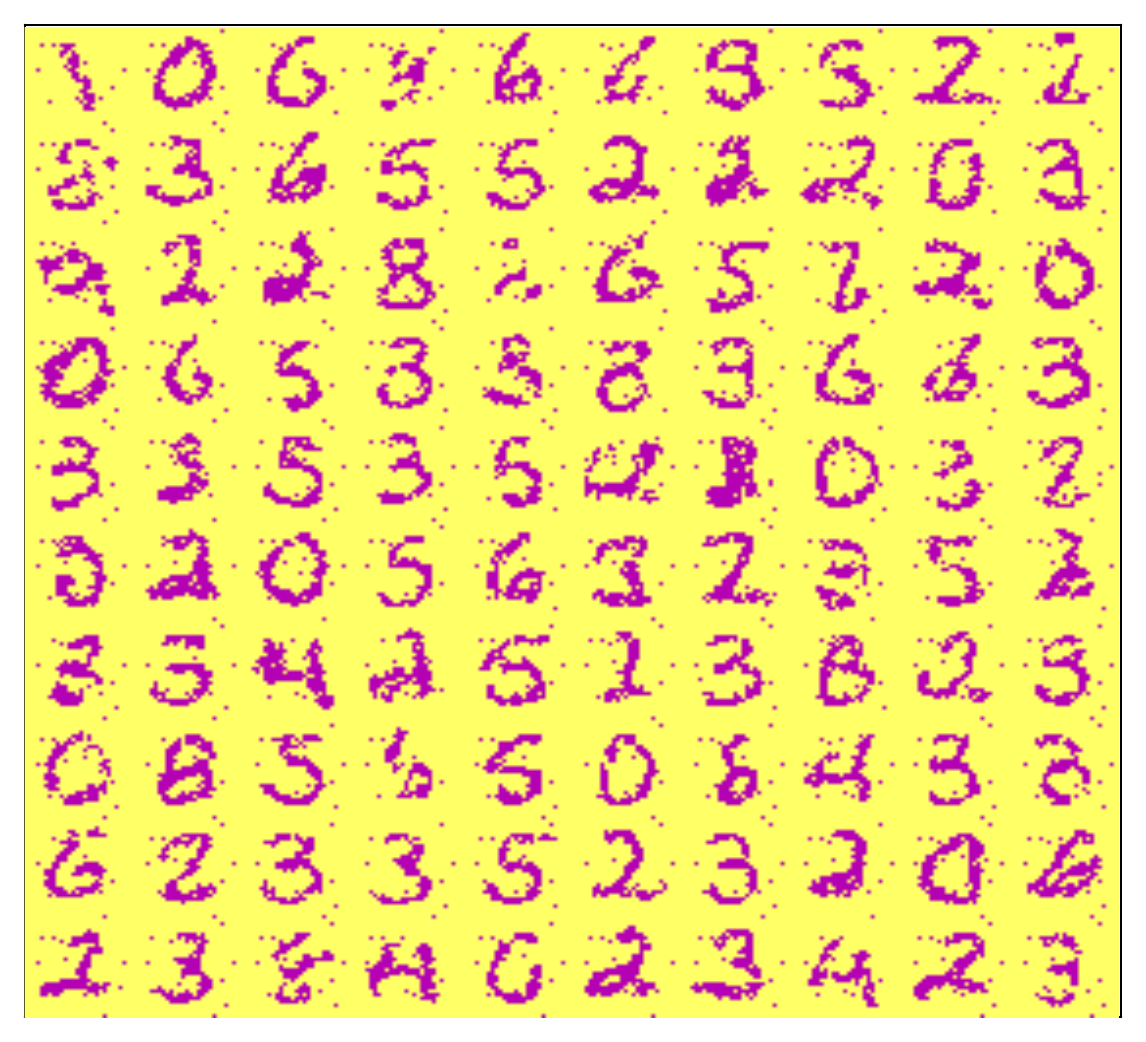}
    \caption{Gibbs sampled data. Same data after rotating by $\pi$ the eigenvector matrix $\bm{U}$ around all but the two dimensions associated with the largest singular values. Same data after 10 consecutive rotations around any $(N-2)$-dimension where the dimensions associated with the five largest singular values are always included and the remaining $N-7$ dimensions are randomly selected. }
    \label{fig:rot}
\end{figure}

\subsection{Hierarchical Learning}
Hierarchical learning is the mechanism by which different pieces of deep learning models learn different levels of abstraction. In the case of deep neural networks, it has been shown that deeper layers in the network learns more complex features \cite{bilal2017convolutional}. In diffusion models, it has been shown that the feature hierarchies occur at different timescale during the denoising process \cite{sclocchi2025phase}. In GANs it has been shown that the class typically clusters around a subset of eigenvectors in the latent space \cite{toledo2023using}. In the case of RBM, it has been shown that during training, the eigenvectors align with the principal components of the dataset \cite{decelle2021restricted}, which we say is similar to hierarchical learning. We can probe this claim by rotating the eigenvector matrix $U$. To do so, notice that when sampling from an RBM, the last step consists on sampling from $\sigma(\bm{W}\bm{h}+\bm{b})=\sigma(\bm{U} \bm{\Sigma} \bm{V}^t \bm{h}+\bm{b})$. Consider a rotation matrix $\mathcal{R}$, and define the rotated orthogonal matrix $\bm{U}_R = \mathcal{R} \bm{U} \mathcal{R}^t$. We build a rotation matrix using the method described in Ref. \cite{aguilera2004general}. This method performs rotations in $n-$dimensions around any arbitrary $(n-2)$-dimension subspace. In Fig. \ref{fig:rot} we show a sample generated by block Gibbs. We also show the effect on the samples of rotating by $\pi$ the eigenvector matrix $\bm{U}$ around all but the two dimensions associated with the largest singular values. Similarly, we show the case where we do 10 consecutive rotations around any $(N-2)$-dimension where the dimensions associated with the five largest singular values are always included and the remaining $N-7$ dimensions are randomly selected. Notice how rotating the eigenvectors associated with the largest singular values has a global effect on the image, whereas rotating eigenvectors associated with any but the principal eigenvectors has a local effect on the image. In this sense, the eigenvectors are associated with different feature hierarchies.

In particular, performing a $\pi$-rotation on the principal eigenvector has the effect of flipping many bits together, reminiscent of a ferromagnetic phase transition. In what follows, we introduce a mean field, and show that the energy landscapes associated to non-zero singular values can be approximated by free energy similar to the paramagnetic-ferromagnetic free energy in Landau theory.

\subsection{Symmetry breaking in energy landscape}

\begin{figure}
    \centering
    \includegraphics[width=3.2in]{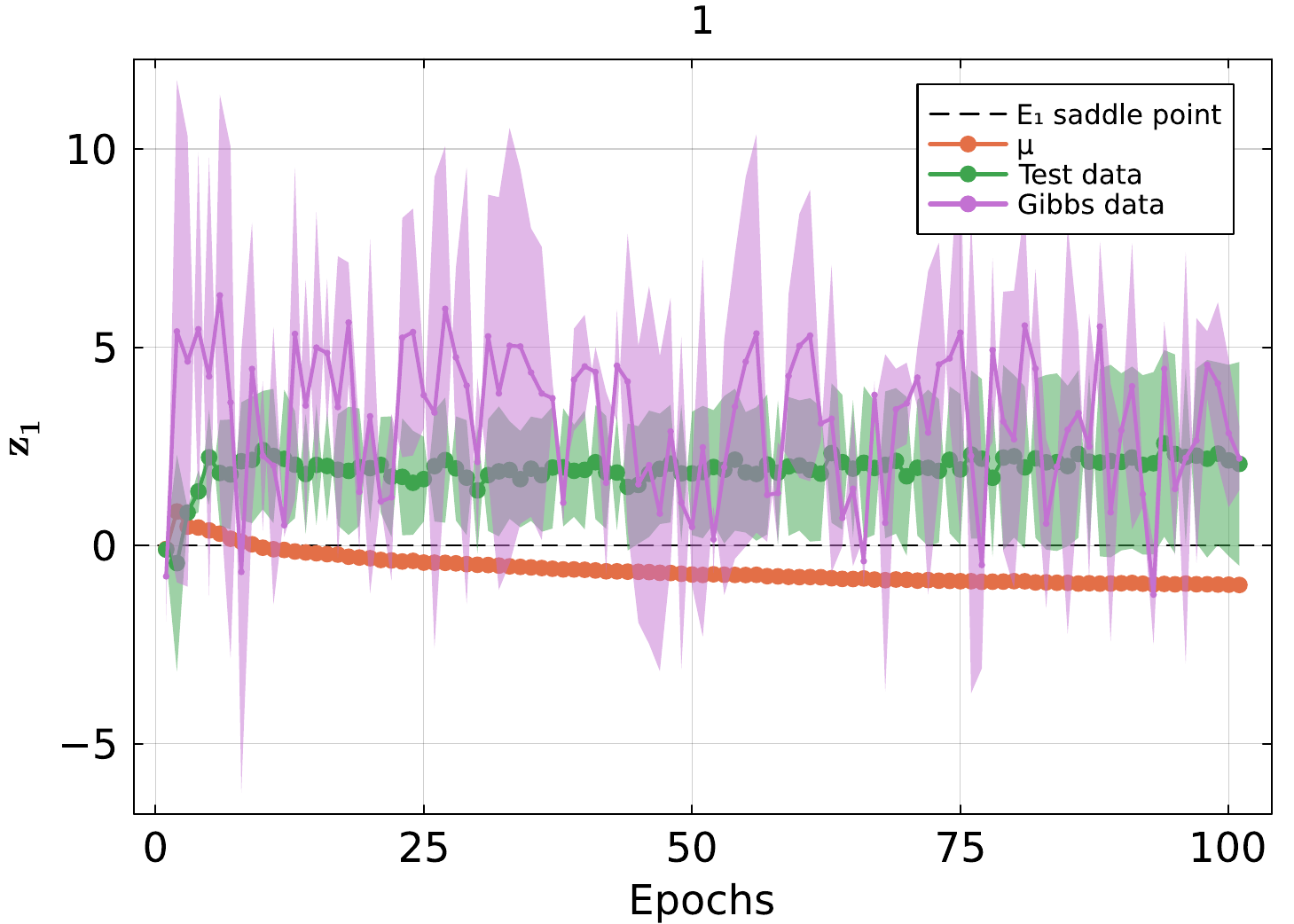}
    \includegraphics[width=3.2in]{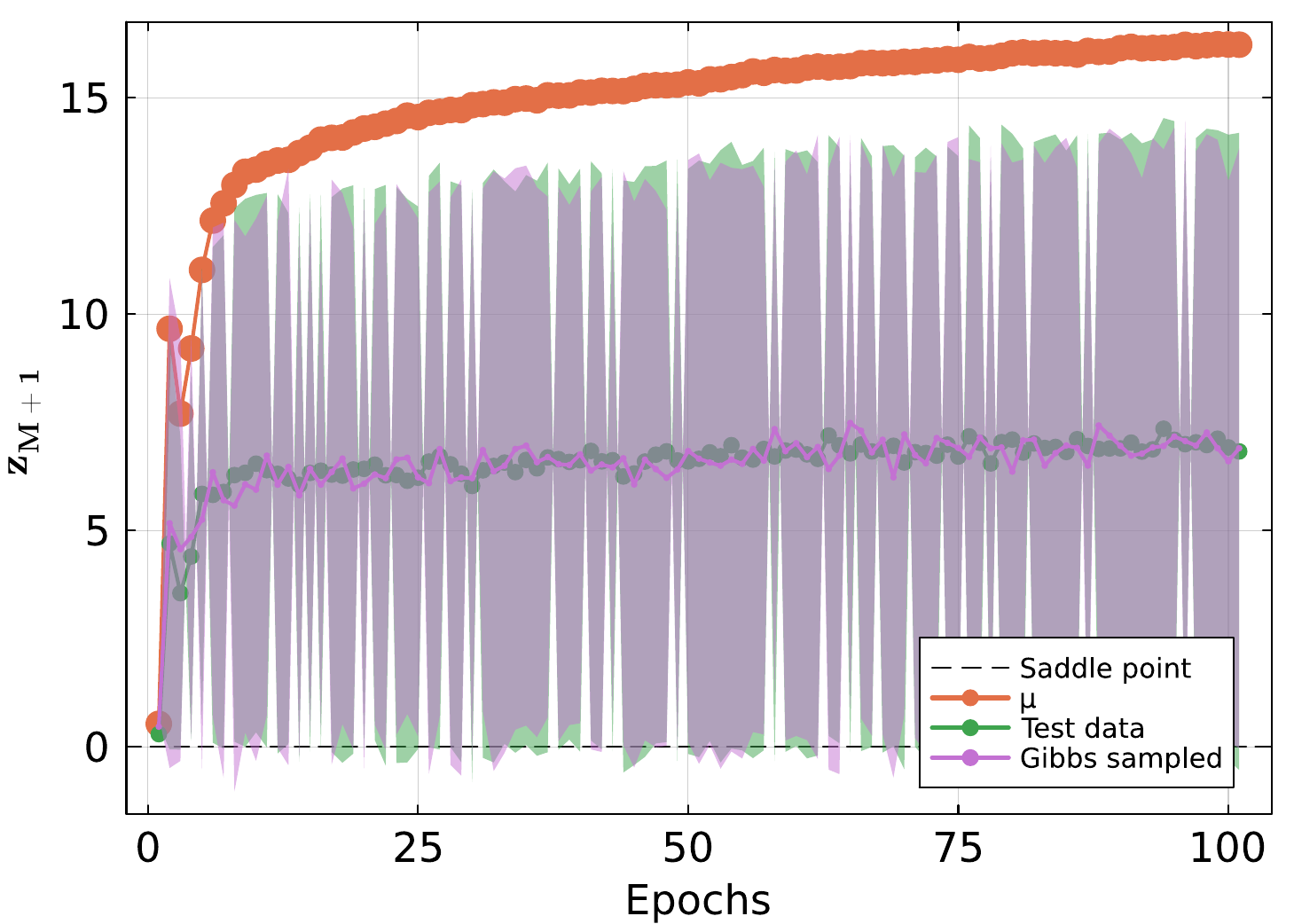}
    \caption{Reciprocal variable $z_1$ (left) and $z_{M+1}$ (right) \textit{vs} epochs for an RBM with $M=500$ hidden nodes. The dashed horizontal line marks the position of the energy function saddle point. The orange markers correspond to the constraint potential minima (see Eq. \eqref{eq:saddlePoint}). The green markers correspond to the average in reciprocal space of the test dataset, whereas the purple markers correspond to the average over reciprocal space of Gibbs sampled data. The ribbons correspond to the standard deviation.}
    \label{fig:enLand_epochs}
\end{figure}

In this section we show how the energy landscape associated with the largest singular value presents a symmetry breaking in the Landau sense. We start from the effective potential presented in Eq. \eqref{eq:effectivepotential}. We can expand the constraint potential around the minimum $\bm{\mu}$ up to fourth order, and after arranging terms we reach:
\begin{eqnarray}
    U_{eff}(\bm{z}) = - \frac{S_c}{\beta} + \frac{V_0}{\beta} + \sum_i U_{eff}^{(i)}(\bm{z})
\end{eqnarray}
with 
\begin{eqnarray}
    U_{eff}^{(i)}(\bm{z}) &=& \frac{a_{0i}b_{0i}}{\lambda_i} + \frac{\Lambda_i}{2} z_i^2 + (z_i - \mu_i) \Gamma_{i}(\lbrace z_p \rbrace_{\substack{p = 1 \\ p \neq i}}^{M+N} ) + (z_i - \mu_i)^2 \Theta_{i}(\lbrace z_p \rbrace_{\substack{p = 1 \\ p \neq i}}^{M+N} ) + (z_i - \mu_i)^3 \Psi_{i}(\lbrace z_p \rbrace_{\substack{p = 1 \\ p \neq i}}^{M+N} ) \nonumber \\
    && + (z_i - \mu_i)^4 \Phi_{i}(\lbrace z_p \rbrace_{\substack{p = 1 \\ p \neq i}}^{M+N} )
\end{eqnarray}
where we have introduced the following functions:
\begin{equation}
    \begin{cases}
        \Gamma_{i}(\lbrace z_p \rbrace_{\substack{p = 1 \\ p \neq i}}^{M+N} ) = \frac{1}{2 \beta}\sum_{j\neq i} k_{ij} (z_j - \mu_j) + \frac{1}{6 \beta} \sum_{\substack{j \neq i \\ k\neq i}} k_{ijk} (z_j - \mu_j) (z_k - \mu_k)  
    \\
    \qquad \qquad \qquad \quad + \frac{1}{24 \beta} \sum_{\substack{j \neq i \\ k\neq i \\ l\neq i}} k_{ijkl} (z_j - \mu_j) (z_k - \mu_k) (z_l - \mu_l)  \\
        \Theta_{i}(\lbrace z_p \rbrace_{\substack{p = 1 \\ p \neq i}}^{M+N} ) =  \frac{k_{ii}}{2\beta} + \frac{1}{6\beta} \sum_{j \neq i} \Pi_{iij} k_{iij} (z_j - \mu_j) + \frac{1 }{24\beta}
     \sum_{\substack{j \neq i \\ k\neq i}} \Pi_{iijk} k_{iijk} (z_j - \mu_j)(z_k - \mu_k) \\
        \Psi_{i}(\lbrace z_p \rbrace_{\substack{p = 1 \\ p \neq i}}^{M+N} ) = \frac{k_{iii}}{6 \beta} + \frac{1}{24 \beta} \sum_{j\neq i} \Pi_{iiij} \\
        \Phi_{i}(\lbrace z_p \rbrace_{\substack{p = 1 \\ p \neq i}}^{M+N} ) = \frac{k_{iiii}}{24\beta} 
    \end{cases} \label{eq:FuncAvgd}
\end{equation}
and $\Lambda_i = \lambda_{i-M} $ for $i>M$ and $\Lambda_i = - \lambda_i $ otherwise.
Notice that the previous functions depend on all $z$-variables but $z_i$. Determining all the parameters in \eqref{eq:FuncAvgd} in general is not tractable. However, we can introduce a mean field, such that $z_k = \mu_k + \epsilon_k$ with $\langle \epsilon_k \rangle = 0$ and $\langle \epsilon_k \epsilon_l \rangle = \delta_{kl}$. We then average the functions in \eqref{eq:FuncAvgd} and obtain:
\begin{equation}
    f_i(z_i) \equiv \langle U_{eff}^{(i)} \rangle (z_i) = \frac{a_{0i}b_{0i}}{\lambda_i} + \frac{\Lambda_i}{2} z_i^2 + \langle \Gamma_i \rangle (z_i - \mu_i) + \langle \Theta_i \rangle (z_i - \mu_i)^2  + \langle \Psi_i \rangle (z_i - \mu_i)^3 + \langle \Phi_i \rangle (z_i - \mu_i)^4
\end{equation}
which resembles the free energy in Landau theory. To validate this approximation, in Fig. \ref{fig:enLand_epochs} we plot the evolution of $\mu_i$ \textit{vs} epochs for $i=1$ and $i=M+1$. We also include the projection of Gibbs sampled data and test data projected onto the energy landscape, which we can assume correspond to the effective potential minimum. For $i=1$ we observe a symmetry breaking occurring at early epochs where the free energy evolves from an harmonic well to a double well, reminiscent of a ferromagnetic phase transition. The fact that $\mu_i$ is located on opposite side of the local minimum with respect to the origin indicate that the double well is non-symmetric. Hence the interplay between the energy function and the constraint potential during training induces symmetry breaking in reciprocal space. Specifically, an increase in magnitude of the prefactor $-\lambda_i/2$ together with a shift of value of the minimum position of the constraint potential, $\mu_i$, from zero to non-zero leads to the symmetry breaking. In contrast, for $i=M+1$ we observe that the corresponding mode becomes relatively flat. 

\begin{figure}
    \centering
    \includegraphics[width=6.2in]{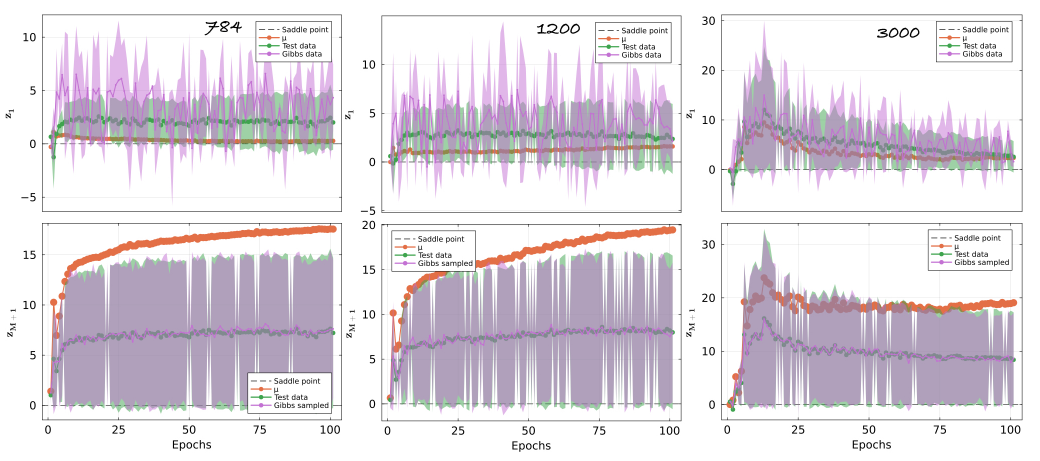}
    \caption{Reciprocal variable $z_1$ (left) and $z_{M+1}$ (right) \textit{vs} epochs for RBMs with (\textbf{left}) $784$, (\textbf{center}) $1200$ and (\textbf{right}) $3000$ hidden nodes. The dashed horizontal line marks the position of the energy function saddle point. The orange markers correspond to the constraint potential minima (see Eq. \eqref{eq:saddlePoint}). The green markers correspond to the average in reciprocal space of the test dataset, whereas the purple markers correspond to the average over reciprocal space of Gibbs sampled data. The ribbons correspond to the standard deviation.}
    \label{fig:enLand_epochsMore}
\end{figure}

This characteristic persists across different hidden layer sizes, as illustrated in Fig. \ref{fig:enLand_epochsMore}. As the hidden layer size increases, the constraint potential minimum, $\mu_i$, shifts closer to the effective potential minimum obtained by the test set and block Gibbs sampling. In general, the symmetry breaking is explicit, since the free energy is asymmetric. In the special case where the number of hidden nodes equals the number of visible nodes, the constraint potential minimum tends to align closely with the energy function saddle point. However, the free energy is still asymmetric, in general, due to the first- and third-order terms in the free energy. \revthree{This result is in agreement with Ref. \cite{decelle2018thermodynamics}, where the authors also identified a mechanism of symmetry breaking by studying the equilibrium properties of RBMs, and further showed the transition from a paramagnetic phase to a ferromagnetic phase driven by the learning dynamics.}

\section{Discussion}
\revthree{In the preceding sections we established a clear connection between RBMs and diffusion process, but a direct link with diffusion models is pending. Such a connection could have a great impact in Quantum Machine Learning, as it could enable the use of quantum annealers to implement diffusion models with faster sampling. The main obstacle is that diffusion models, in general, do not satisfy the detailed balance condition. Nonetheless, different approaches have been proposed. For instance, in Ref. \cite{ambrogioni2025statistical} the author proposes an adiabatic diffusion model, while in Ref. \cite{song2023consistency} the authors introduce consistency models whereby a separate neural network is trained to mimic the diffusion model, enabling one-step sample.} 
In addition, we established a clear and novel connection between RBMs and coupled Bosons, which may provide new pathways that relate RBM with problems in many-body physics. 
\revthree{We showed that sampling the weight matrix from a Gaussian distribution renders the RBM weight distribution rotationally symmetric. For an infinite size hidden layer, the reciprocal variables associated with the non-zero singular values are Gaussian distributed. In this limit, the associated Bosonic system decouples into a set of uncoupled harmonic oscillators, with eigenvalues $E_i=4\pm \lambda_i \beta$. There is a critical singular value, $\lambda_c=4/\beta$, for when $\lambda>\lambda_c$, the bound states hinder the diffusion process from reaching the Boltzmann distribution. Consequently, RBM configuration can become trapped in non-Boltzmann states, leading to biased gradient estimates during maximum-likelihood training. It is possible, however, that after a few training steps, once this symmetry is broken the reciprocal variables are no longer Gaussian distributed and the coupled Bosons bound states disappear. Testing this is left for future work.}
    
\revone{Our results also provide a theoretical understanding for known rules-of-thumb in RBM which renders the RBM rotational symmetry and provided:
    \begin{itemize} 
        \item RBM layer sizes and weights variance. The size of the hidden layer typically is less or equal to the visible layer. Our results show that for a large difference between hidden and visible layer sizes and/or weights with large variance, the energy landscape can be unstable at initialization.

        \item Instabilities in Gaussian RBMs. For Gaussian‐Gaussian RBMs, the energy function is effectively a set of hyperbolic paraboloids in reciprocal space, while the corresponding constraint potential is relatively flat. This allows the reciprocal variables to drift, rendering Gaussian RBMs especially prone to instability and difficult to train.

        \item Regularizers. The use of regularizers during training is a common approach when training generative models. Particularly, the L2-norm is used when training an RBM. In maximum posterior training, one typically maximizes
        \begin{equation}
            \log P(\text{weights}|\text{data}) = \log P(\text{data}|\text{weights}) + \log P(\text{weights}) - P(\text{data}) \; .
        \end{equation}
        where the term $\log P(weights)$ is the L2-norm, and can be understood as a constraint on the model enforcing rotational symmetry in the RBM. However, we note that breaking this symmetry is important for hierarchical learning, but it does not guarantee that the model captures the data set distribution. In some cases, the symmetry may break without improving performance. A more tailored L2-norm, where the constraints are applied with different weights on linearly independent directions of the weight matrix can encourage symmetry breaking in a way that enhances the learning.
    \end{itemize}
    }
\revthree{Rotational symmetry in RBMs can be viewed as analogous to starting a sculpture from a perfectly symmetrical block of material. Training corresponds to carving away symmetries until the model captures the desired data structure. This underscores the relevance of rotational symmetry in RBMs. However, as we stressed before, under certain conditions this symmetry may hinder the RBM training. In Ref. \cite{hou2020statistical} the authors show a symmetry breaking from a phase the \textit{random guess} phase where there is no symmetry breaking and the data do not provide any useful information to bias the weight’s direction to a phase where the reverse symmetry breaks. In Ref. \cite{bachtis2024cascade} the authors show that RBMs undergo a second order phase transition during training at early epochs. It remains to be unvelied the full phase diagram in RBMs. We leave this exciting experiment for future work, as this will require intensive numerical results.}

\section{Conclusions} \label{sec:discussion}
In the previous sections we introduced a reciprocal space method. This method allows a better understanding on RBM initialization and training. With this method we showed a direct mapping from a Restricted Boltzmann Machine to a diffusion process. This diffusion is governed by an effective potential that encapsulates the discrete nature of the binary variables through inherent constraints. Although the mapping is quite general, its practical implications have yet to be fully explored.  Furthermore, we established a direct connection between an RBM and a many-body problem, specifically, a system of coupled Bosons, in which the eigenvalues are proportional to the relaxation times in the RBM and the logarithm of the ground state is proportional to the effective potential. We also showed that the reciprocal variables are Gaussian distributed to first approximation. As the number of hidden nodes grows, the reciprocal variables tend to Gaussianity. Consequently, if the reciprocal variables are Gaussian, the diffusion process diverges from the stationary solution for certain modes. We intend to investigate this prediction in future work. Another interesting result relates to the distribution of singular values in the case of trained RBMs, shown in Fig. \ref{fig:rhoLambda}, where despite the different sizes considered for the hidden layer, the distribution is qualitatively similar. We systematically observed a gap at $\lambda \in [33,35]$ for trained RBMs and currently we have no explanation for this.

We further demonstrated that training an RBM induces a break in rotational symmetry, which in turn affects the learning process by enabling the weight matrix eigenvectors to capture different data features in a hierarchical sort. In particular, rotating the eigenvectors associated with the largest singular value has a non-local effect on the binary space. Within the framework of statistical physics, we can interpret the reciprocal variable associated with the largest singular values as an order parameter. Moreover, we showed that the effective potential associated with non-zero singular values can be expressed in terms of a free energy formulation reminiscent of Landau theory. Finally, our results reveal a symmetry breaking in the free energy landscape during training. It remains to be seen whether increasing the number of hidden nodes hinders the symmetry breaking and how this impacts the learning process.


\section{Acknowledgements}
JQTM is thankful to Jorge Fernandez de Cossio, Daniel Miravet, Mehdi Drissi and Ejaaz Merali for useful discussions. The authors thank the reviewers for their comments and discussion. JQTM acknowledges a Mitacs Elevate Postdoctoral Fellowship (IT39533) with Perimeter Institute for Theoretical Physics.
We gratefully acknowledge funding from the National Research Council (Canada) via Agreement AQC-002.
This research was supported in part by Perimeter Institute for Theoretical Physics. Research at Perimeter Institute is supported by the Government of Canada through the Department of Innovation, Science and Economic Development and by the Province of Ontario through the Ministry of Research, Innovation and Science. The University of Virginia acknowledges support from NSF 2212550 OAC Core: Smart Surrogates for High Performance Scientific Simulations and DE-SC0023452: FAIR Surrogate Benchmarks Supporting AI and Simulation Research. This research was supported in part by grants NSF PHY-1748958 and PHY-2309135 to the Kavli Institute for Theoretical Physics (KITP).

\bibliographystyle{unsrt}  
\bibliography{references}

\newpage

\appendix
\counterwithin*{equation}{section}
\renewcommand\theequation{\thesection\arabic{equation}}
\setcounter{figure}{0}

\end{document}